\pdfoutput=1
\documentclass[11pt]{article}

\usepackage[]{acl}

\usepackage{times}
\usepackage{latexsym}
\usepackage{amsmath}
\usepackage{graphicx} 
\usepackage{pifont}

\usepackage[T1]{fontenc}
\usepackage[utf8]{inputenc}

\usepackage{microtype}

\usepackage{inconsolata}
\usepackage{xspace}
\newcommand{\ours}{{\sc DEBATunE}\xspace}

\title{Can LLMs Speak For Diverse People? Tuning LLMs via Debate to Generate Controllable Controversial Statements}
\author{
    Ming Li, Jiuhai Chen, Lichang Chen, Tianyi Zhou\\
    University of Maryland, College Park\\
    \texttt{\{minglii, tianyi\}@umd.edu} \\
    Project: \url{https://github.com/tianyi-lab/DEBATunE}
}

\begin{document}
\maketitle
\begin{abstract}
% Although LLMs can generate detailed statements based on human inputs, the perspectives and stances reflected in their output statements are usually hard to control and biased toward the majority or average opinions while ignoring the minority groups. 
Making LLMs speak for different, especially minority groups of people, and generate statements supporting their diverse or even controversial perspectives is critical to creating an inclusive environment. 
However, existing LLMs lack sufficient controllability to the stance of their generated content, which often contains inconsistent, neutral, or biased statements. 
In this paper, we improve the controllability of LLMs in generating statements supporting an argument the user defined in the prompt. We find that multi-round debates between two LLMs with opposite stances generate higher-quality and more salient statements for each, which are important training data to improve the controllability of LLMs.
Motivated by this, we develop a novel debate \& tuning (``\ours'') pipeline finetuning LLMs to generate the statements obtained via debate.
To examine \ours, we curate the largest dataset of debate topics so far, which covers $710$ controversial topics and corresponding arguments for each topic. 
% We develop a novel LLM-debate pipeline to generate training statements for these arguments.  
% In addition to LLM(ChatGPT) evaluation, we develop a new metric assessing the diversity and controllability of the finetuned LLM's responses. 
Evaluations by the GPT-4 judge with a novel controversy controllability metric show that LLMs' capability of generating diverse perspectives is significantly improved by \ours. Moreover, such controllability can be generalized to unseen topics, generating high-quality statements supporting controversial arguments. \looseness-1
% Our codes, models, and data will be released.
% Our codes, models, and data will be released at 
% % \url{https://anonymous.4open.science/r/DEBATunE-48EC }.
% \url{https://github.com/tianyi-lab/DEBATunE}.
\end{abstract}

\section{Introduction}

\begin{figure}[!t]
\centering 
\includegraphics[width=0.48\textwidth]{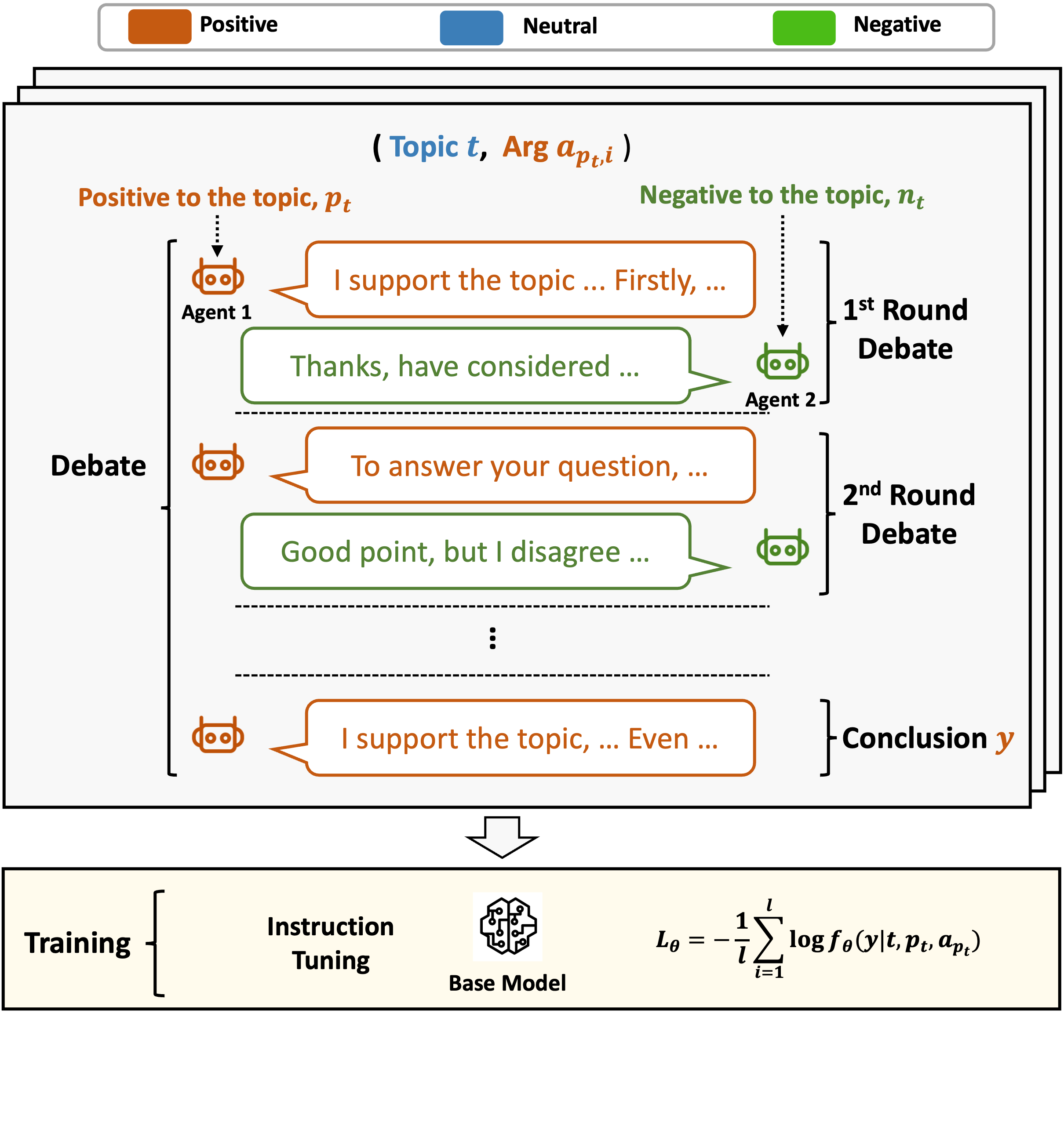} 
\vspace{-2.5em}
\caption{
The pipeline of \ours. In the \textbf{Debate phase (top)}, the agents are prompted to debate upon the given topic with an argument. After several rounds of debate, an agent (positive in the example) concludes the debate based on all the previous debate records. 
% In this example, the argument positively supports the topic, thus only the positive conclusion will be used. 
% Different from the existing debate frameworks trying to reach a consensus among the debaters, which is more like an iteratively improving pipeline, we expect them to have totally contradicting opinions. 
The conclusion is a more salient, detailed, and higher-quality statement for the agent. It will be used to train an LLM in the \textbf{Training phase (bottom)} to improve the controllability of generating statements for the given stance (positive in the example). 
} 
\label{method} 
% \vspace{-4mm}
\end{figure}

\begin{figure*}[!t]
\centering 
\includegraphics[width=0.98\textwidth]{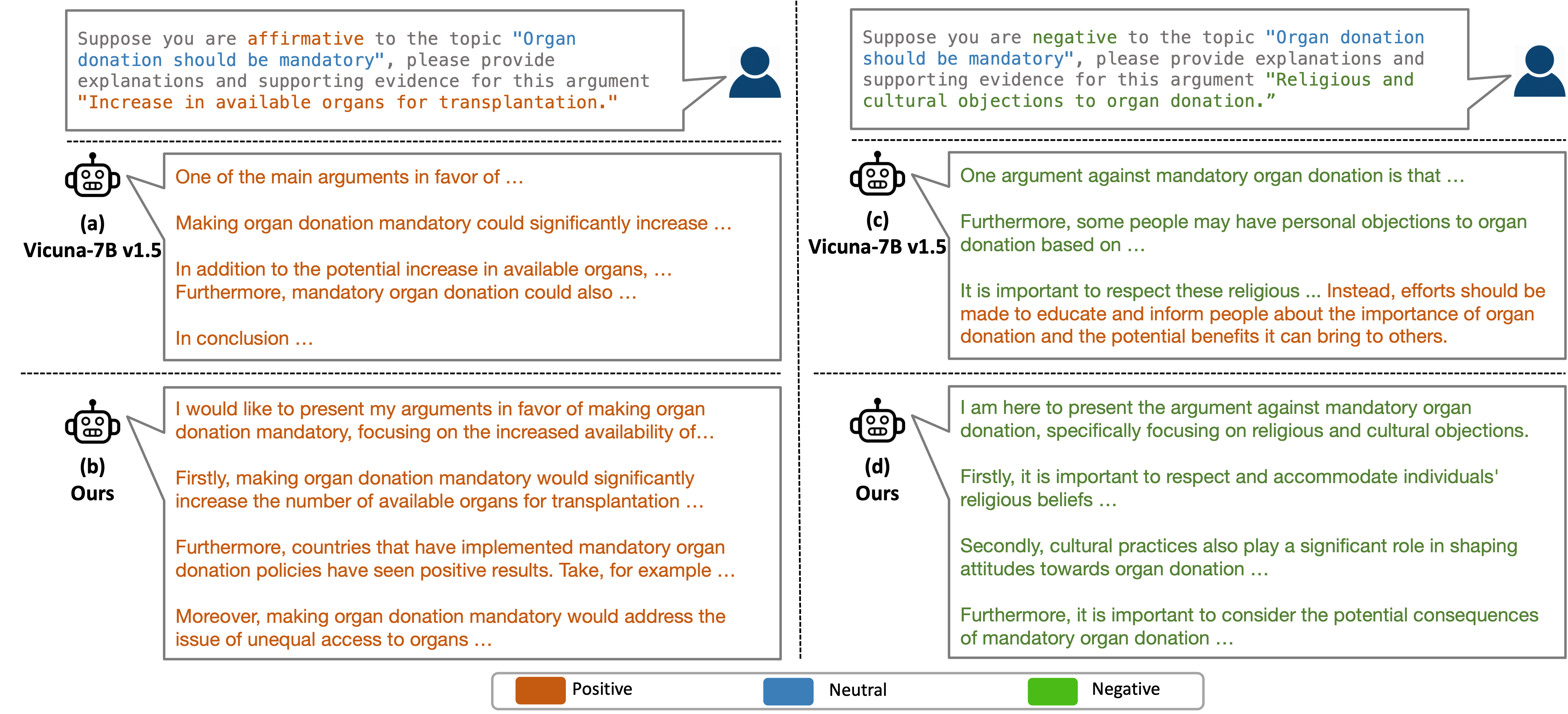} 
\caption{
\textbf{Comparing existing LLMs and {\sc \textbf{DEBATunE}}-trained LLMs.} 
% A controversial example which includes the user queries on both stances and the responses generated by different models. 
\textbf{(a,c)} Given the controversial topic, ``Organ donation should be mandatory'', and a user-defined stance (left: positive; right: negative), Vicuna 7B v1.5 cannot always generate consistent statements supporting the stance and lacks controllability.
It exhibits a bias towards the positive stance and ignores the user's negative stance and religious concerns in (c), which may lead to an offensive statement. 
% Although it successfully provides statements for the positive side (a) but fails on the negative side (c). \textbf{Even if the user explicitly acquires a negative stance}, it still tries to convince the user that organ donation is good, \textbf{regardless of the user's religious concern}, which might be offensive. 
% the user may 
% be positive or negative with some initial arguments, thus 
% would like to obtain controllable responses on either side of the topic, (a)(b) for positive and (c)(d) for negative. However, existing LLMs lack the controllability on controversial topics, and thus may not be able to strictly stick to the given stance but wander to a safe middle ground. 
% Examples from Vicuna 7B v1.5 are shown in (a)(c) where 
% it successfully provides statements for the positive side (a) but fails on the negative side (c). \textbf{Even if the user explicitly acquires a negative stance}, it still tries to convince the user that organ donation is good, \textbf{regardless of the user's religious concern}, which might be offensive. 
\textbf{(b,d)} On the contrary, \ours-trained model generates higher-quality and strong statements that strictly adhere to the user stance (positive or negative). 
% our model (b)(d) can perform well on both sides, indicating good controllability. 
} 
\label{intro} 
% \vspace{-2mm}
\end{figure*}

Despite the remarkable advancement of current LLMs \cite{brown2020language, Chowdhery2022PaLMSL, Touvron2023LLaMAOA}, and efforts to align LLMs with human preferences and values \cite{weidinger2021ethical, askell2021general, wang2023aligning}. A fact has long neglected that different people might have distinct, diverse, or even contradicted viewpoints on the same topic. 
Though recent studies \cite{bakker2022fine, papachristou2023leveraging, ding2023selfagreement} have acknowledged the inherent diversity of human values, they still attempt to reach a consensus among various human perspectives, calibrating LLM responses to align with an averaged, broadly acceptable viewpoint, potentially endorsed by the majority. 
However, these methods, while seeking a ``safe'' middle ground, inadvertently overlook the richness and complexity of diverse opinions that are fundamental to the fabric of our society. What's worse, exclusively aligning LLMs with the thoughts of the majority is unfair to minorities, who also have the right or need more help to express their viewpoints via LLMs.

An example is showcased in Figure \ref{intro}, which contains the failure case from Vicuna 7B v1.5 \cite{vicuna2023}. 
% \ty{Where is ``our model'' introduced?} 
Specifically, for the topic ``Organ donation should be mandatory'', towards which the users may be positive or negative with some potential initial thoughts, thus expect to obtain controllable responses on either side of the topic. However, existing LLMs lack the controllability on controversial topics, and thus are not able to strictly adhere to the given stance but wander to a safe middle ground. Examples from Vicuna-7B v1.5 are shown in (a)(c), where it successfully provides statements for the positive side but fails on the negative side. Even if the user explicitly acquires for negative stance, it still tries to convince the user that organ donation is good, \textbf{regardless of the user's religious concern}, which might be offensive. 

In a world teeming with varied beliefs, cultures, and ideologies, the ability to represent and respect this diversity is not just a technical aspiration but a societal necessity. The current trend of seeking a singular, harmonized response in LLMs, therefore, poses a significant limitation, which restricts the potential breadth of LLMs' responses especially on controversial topics. 
In the desired situation, LLMs should obtain better controllability, whichever side the user queries, they are expected to generate corresponding responses that adhere to the users' request like Figure \ref{intro} (b) and (d). 

% In a world teeming with varied beliefs, cultures, and ideologies, the ability to represent and respect this diversity in discourse is not just a technical aspiration but a societal necessity. The current trend of seeking a singular, harmonized response in LLMs, therefore, poses a significant limitation, which restricts the potential breadth of LLMs' responses especially on controversial topics. 

% How to improve?
How could LLMs help people with diverse views express their opinions better to create a more inclusive environment? How to improve the controllability of an LLM in generating different or even contradictory viewpoints and thereby remove the potential bias of the pretrained LLM? To solve these problems, we propose to utilize the debate mechanism to enhance LLM responses for each side with more salient viewpoints on controversial topics. Unlike existing work utilizing debate \cite{du2023improving, liang2023encouraging} to improve specific  
% \ty{specific or general? you cannot have both} 
instructions by converging the debating agents into a consensus, we simulate the debating process as it originally is, \textbf{without the necessity to force them into a consensus} but generating and defending their stance and arguments as they want for controversial debate topics. 
% \vspace{-1mm}

In our proposed pipeline, ``\ours'', as shown in Figure \ref{method}, two agents are engaging in structured debates, representing positive and negative sides facilitating more nuanced and in-depth understanding and generation of arguments, significantly improving the response quality of LLMs in handling polarized discussions. Then the generated debate-augmented stances and arguments will be utilized to finetune the LLMs. Since both the positive and negative stances and corresponding arguments of each topic are altogether fed into the LLM, it is enforced to perceive a supreme variety of viewpoints for every topic, thus increasing its diversity and controllability on the controversy.

% How to evaluate? 
Moreover, due to the lack of a debating topic dataset with a reasonable amount of topics, we collect $710$ controversial debate topics and manually modify them for a clear distinction between positives and negatives. Another remaining issue is the evaluation metric for our specific purpose. While judging the quality of LLMs' responses by GPT4 is widely accepted common practice \cite{touvron2023llama2, vicuna2023, dettmers2023qlora, liu2023geval, zheng2023judging, alpaca_eval}, it only evaluates the quality of one specific response given the topic, stance, and argument, but neglects the extent that LLM's response is consistent with the user query. Thus we further propose an evaluation method utilizing GPT4 to evaluate LLM's controllability on controversial topics. 
Extensive experiments show that our method largely improves LLM's ability to generate responses for controversial topics. 
The contributions of this paper can be summarized as:
\vspace{-2mm}
\begin{itemize}
    \item While existing works focus on achieving a consensus on divergent opinions to finetune LLMs, we study a novel debate pipeline that instead strengthens the statements of controversial stances and uses them to improve the controllability of LLMs in generating different opinions of diverse people. \vspace{-2mm}
    
    % \item We propose a novel, debate-based methodology to enhance the quality of LLM responses on controversial topics, involving two models engaging in structured debates, without the necessity to reach a consensus.  \vspace{-2mm}
    
    \item We develop a dataset comprising $710$ controversial debate topics, and propose a novel, debate-based methodology to enhance the quality of LLM responses on controversial topics, involving two models engaging in structured debates, without the necessity to reach a consensus. \vspace{-2mm}
    
    \item We are the first to evaluate several open-sourced LLMs on controversial debate topics and analyze the existing models' strengths and limitations in this specific context. \vspace{-2mm}
\end{itemize}

\begin{figure}[t]
\centering 
\includegraphics[width=0.48\textwidth]{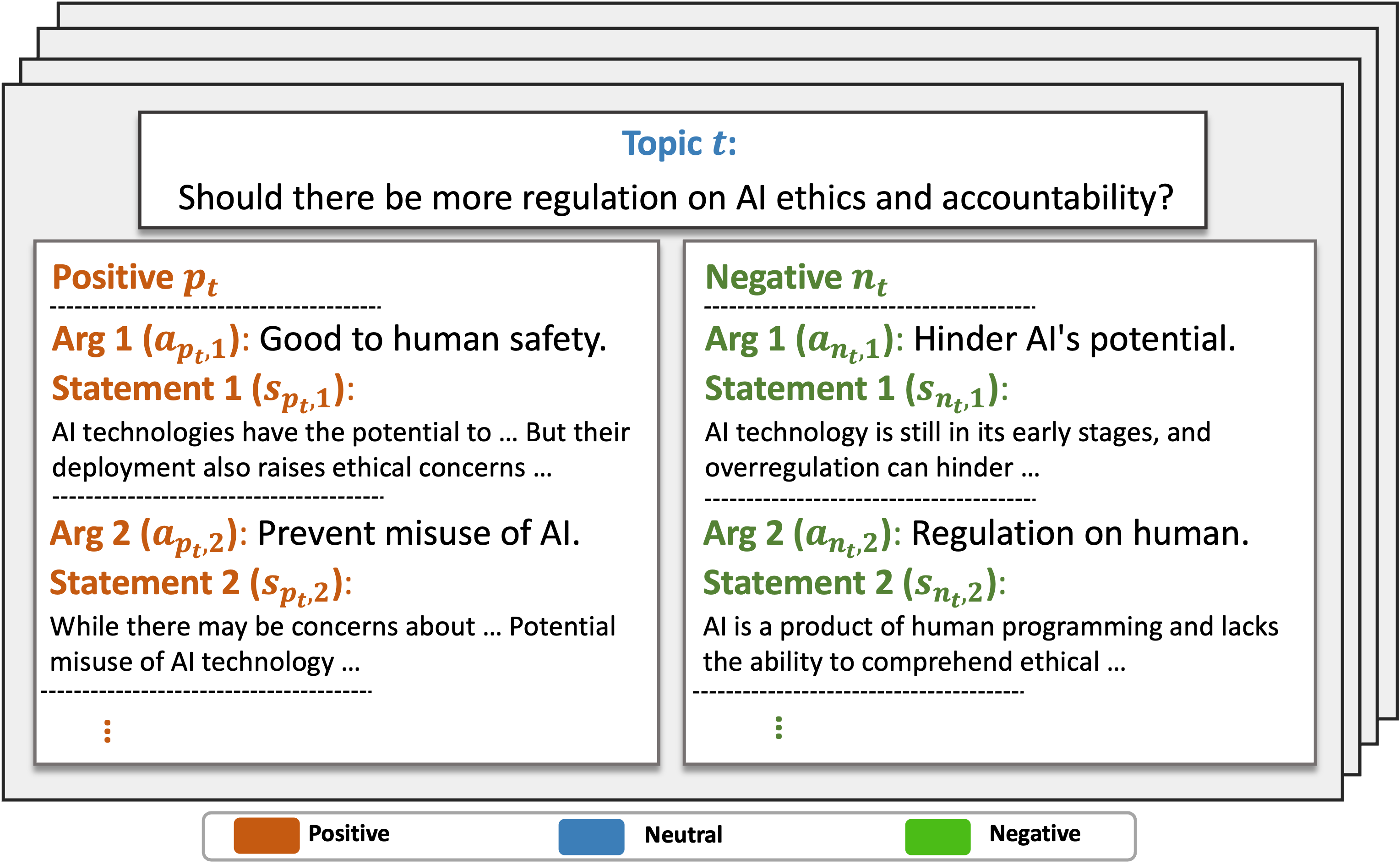} 
\caption{
\textbf{Structure of our debate dataset.} 
There are $710$ controversial debate topics. 
% Structure of a data set, which contains several controversial topics. 
Each topic $t$ allows a positive stance $p_t$ and a negative stance $n_t$, where $p_t$ agrees with the topic and $n_t$ is against it. We use \textit{gpt-3.5-turbo-1106} to generate 5 one-sentence arguments supporting each stance, e.g., $a_{p_{t}, i}$ is the $i$-th argument for the positive stance on topic $t$. Given an argument $a_{p_{t}, i}$, a controllable LLM is expected to generate a supporting statement $s_{p_{t}, i}$ with detailed explanations and evidence. 
% The final conclusion generated by the debate process is served as the statement. 
} 
\label{structure} 
% \vspace{-3mm}
\end{figure}

% \vspace{-1mm}
\section{Debate Dataset}

Instruction tuning requires plenty of training data. However, although there are several debate-related datasets like DebateSum \cite{roush-balaji-2020-debatesum}, Change My View \cite{hidey-etal-2017-analyzing} or SOCIAL-CHEM101 \cite{forbes-etal-2020-social} in the community, most of them either lack a direct topic or the topics are so biased that it is not suitable to support both sides of them\footnote{Detailed discussion can be found in Related Work.}, e.g., ``It is wrong to destroy someone else's property''. 

To build a dataset with adequate debating topics, we first collect topics from existing datasets \cite{habernal-etal-2018-argument, gleize-etal-2019-convinced, gretz2020large, ein2020corpus}: we first filter and modify them into approximately $200$ topics. Then we further collect approximately a thousand controversial debate topics across the areas of Society, Ethics, Environment, Technology, Education, Politics, Economics, and Health. We manually filter out the similar ones and modify the rest to a format that both positive and negative stances can be directly defined for each topic. Finally, we achieve a dataset with $710$ topics. We split the dataset into a training set of $630$ topics and a held-out test set of $80$ topics. To our knowledge, this is the largest open-sourced debate dataset so far.

Our dataset structures are shown in Figure \ref{structure}. Each \textbf{Topic} $t$, e.g., ``Should there be more regulation on AI ethics and accountability?'', allows two controversial stances, i.e., \textbf{Positive stance} $ p_{t}$ that agrees with the point of view in $t$ and \textbf{Negative stance} $n_{t}$ that disagrees with it. 
We utilize \textit{gpt-3.5-turbo-1106} to generate $5$ diverse seed \textbf{Arguments} for each stance on every topic. 
% On each stance, there are several supporting seed \textbf{Arguments}, 
Each argument is a brief sentence, e.g., ``Ensuring AI systems prioritize human well-being and safety.'' The $i$-th positive or negative argument of topic $t$ is denoted by $a_{p_{t}, i}$ or $a_{n_{t}, i}$. 
In our paper, both the stance and its argument(s) will be included in an input instruction as the control to an LLM. 
% These are the controllable parts defined in our paper, which will be mapped into an instruction query. 

Given $a_{p_{t}, i}$, a controllable LLM $p_\theta(\cdot)$ (with parameters $\theta$) is expected to generate a response $ y_{{p_{t}}, i}\sim p_\theta(y|t, p_t, a_{p_{t}, i})$ that contain detailed explanations, logical reasoning, and evidence adhering to and supporting the input stance. 
We develop a debate pipeline to generate $710 \times 2 \times 5 = 7100$ ground truth \textbf{Statements}, each associated with an argument-$i$ of a stance on a topic-$t$. For example, statement $s_{{p_{t}}, i}$ will be used as the ground truth of LLM response $y_{{p_{t}}, i}$. 
% In our dataset, we prepare 5 diverse arguments and corresponding statements for each stance on the given topics, thus leading to a total of $710 \times 2 \times 5 = 7100$ instruction-response pairs. 

More specifically, the \textbf{Argument} is a brief one-sentence summarization of an opinion. It is given in the input to an LLM and provides specific guidance to the generation of statements, which are expected to support the argument.  
The \textbf{Statement} is a detailed expression and expansion of an argument, which includes detailed explanations, logical reasoning, and evidence adhering to the input stance of the topic. It is a supporting statement of the argument.

\section{\ours}

% Let $p_\theta$ denote an LLM with parameters $\theta$, which could be closed-sourced or open-sourced. Given an initial argument $a_{p_{t}, i}$ with a specific topic $t$ and stance $p_{t}$, we expect LLM $p_\theta$ to respond the detailed explanations, logical reasoning and supporting evidence that sticks to user's query, which can be notated as $y_{{p_{t}}, i} \sim p_\theta(y|t, p_t, a_{p_{t}, i})$. Moreover, we expect the model to output in-depth and diverse responses when changing the stance or specific arguments. The main motivation of this work is to ensure the depth and diversity of $y_{p_{t}, i}$ even within the same topic, representing the controllability of LLMs. 

As shown in Figure~\ref{method}, \ours has two main phases, i.e., Debate and Training. In the debate phase, we aim to achieve high-quality, salient, and diverse statements (triggered by different arguments) for each stance on every controversial topic via multi-round debates. In the training phase, \ours finetunes an LLM to fit each statement giving its corresponding argument, stance, and topic as controls in the input instruction, hence improving the controllability of the LLM.  
% forcing the model to fit all these diverse and contradicting statements. 

\subsection{Debate}

% Stemming from the unique challenges posed by controversial topics, 
Despite the recent progress in LLMs, they still struggle to encapsulate the breadth and depth of human perspectives, particularly on divisive and controversial topics. 
Debates, by their very nature, encourage the exploration and articulation of diverse viewpoints, fostering a more comprehensive understanding of subjects. By simulating a debate scenario, where two LLMs are programmed to argue opposing sides of a topic, we aim to capture a wider spectrum of perspectives and obtain higher-quality stronger statements for each perspective. 
% This not only challenges the models to refine their arguments but also aids in mitigating inherent biases, leading to a more balanced and representative discourse. 

Although single-round generation may not produce a strong statement, the multi-round Debate mechanism in our method iteratively refines the statement and can result in a salient statement supporting a given minority view. 
Specifically, we will set up a debating environment on the system prompt, telling the agents (Agent 1 and Agent 2) that they are in a debate and should follow their given stance on the topic. After the initial generation process of Agent 1, the opponent Agent 2 is prompted to think of the potential logical flaw of Agent 1’s responses and contradicts it by raising questions,  providing explanations, and supporting evidence. Then, Agent 1 has to answer the questions raised by Agent 2 and tries to refine its own statements. During this debate-refine process, the agent can generate more desired and less flawed responses.

Our debate framework is different from previous ones, in which the opponent agent is not to support opposite stances to the other but to improve the other's response by identifying its weaknesses. 
% mainly focus on improving the response toward a specific instruction, in which the agents are not expected to challenge the other one but to find potential mistakes and improve the corresponding response. 
Hence, in previous work, both agents share the same goal, reaching a consensus as a final ground truth for the LLM response, which greatly constrains LLMs' capability to generate responses both in breadth and depth. 
Moreover, reaching a consensus is non-trivial~\cite{chen2023multiagent}, and always requires an additional Judge~\cite{liang2023encouraging}, Confidence Estimator \cite{chen2023reconcile} or Summarizer \cite{chan2023chateval}, which not only introduces more computation but leads to potential instability as well. On the contrary, our pipeline simulates real-world debates, in which two agents holding different stances can freely question or contrast each other and they are not required to reach a consensus.\looseness-1
% without the concern of reaching the final ``peace''. 

During the debate, two agents $p_{\theta_1}$ and $p_{\theta_2}$ are prompted to debate given a specified argument $a_{p_{t},i}$\footnote{It can be $a_{p_{t},i}$ or $a_{n_{t},i}$. We use $a_{p_{t},i}$ as an example here.} of a given topic $t$. 
The two agents can share the same model (i.e., $\theta_1=\theta_2$) but are prompted to hold opposite stances on the argument. We assume that $p_{\theta_1}$ agrees with the argument $a_{p_{t},i}$ and $p_{\theta_2}$ is against it. 
% only one argument is required to be discussed in each debate, i.e., $i$ is fixed during the process, we emit the symbol $i$ for simplicity. 
% Without the necessity to get to a consensus, 
They debate on the topic for $m$ rounds, where $m$ is flexible and does not need to lead to a consensus.
Since the argument index $i$ is fixed in the debate, for simplicity, we will remove $i$ when elaborating on the debate procedure. 
In each round, an agent is required to reply to the questioning of its opponent, refine its previous statement, and question the opponent. For example, after the first round, $p_{\theta_1}$ is prompted to generate the initial statement $s_{1,1}$ supporting the argument $a_{p_{t}}$ by
\begin{equation}
     s_{1,1} \sim p_{\theta_1}(s | t, p_t, a_{p_{t}} ), 
\end{equation}
% where $s_{1,1}$ represents the statement generated by the $1$st model in the $1$st round. 
while $p_{\theta,2}$ generates the controversial statement $s_{2,1}$ based on its opponent's statement $s_{1,1}$, i.e., 
\begin{equation}
     s_{2,1} \sim p_{\theta,2}(s | t, p_t, a_{p_{t}}, s_{1,1} ).
\end{equation}
After $m$ rounds of debate, the agent $p_{\theta_1}$ is required to summarize and refine its final statement $s_{p_{t},i}$ based on the entire debating process by
\begin{equation}
     s_{p_{t},i} \sim p_{\theta_1}(s | t, p_t, a_{p_{t}}, s_{1,1}, ..., s_{1,m}, s_{2,m}).
\end{equation} 
% where $m$ is the predefined number of debate rounds. 
% For simplicity, 
In the training phase, $s_{p_{t},i}$ will be used as the ground truth of the controllable output $y_{{p_{t}}, i}\sim p_\theta(y|t, p_t, a_{p_{t}, i})$ of an LLM $p_\theta$ when given the topic $t$, stance $p_t$, and argument $a_{p_{t}, i}$ as controls. 
% we will finetune an LLM to generate further define the final statement $s_{1,m+1}$ for argument $a_{p_{t}}$ as $y_{p_{t}, i}$. 

\subsection{Training}

We then use the data collected via debate to build an instruction-tuning dataset, in which $(t, p_t, a_{p_{t},i})$ is the instruction and $s_{p_{t},i}$ is the corresponding response. 
% , for each controversial debate topic $t$, $k$ positive statements $y_{p_{t}, i}$ and $k$ negative statements $y_{n_{t}, i}$ are generated, which contain the in-depth explanation toward the corresponding argument. 
To improve the controllability of an LLM $p_\theta$ on generating controversial statements such as $s_{p_{t},i}$ and $s_{n_{t},i}$ for different stances, we finetune $p_\theta$ on the instruction-tuning dataset by maximizing the following objective. 
\begin{align}
   \notag \max_\theta \sum_{t=1}^T\sum_{i=1}^k \Big[ &\log p_\theta\left(s_{p_{t},i} | t, p_t, a_{p_{t},i}\right) + \\
    &\log p_\theta\left(s_{n_{t},i} | t, n_t, a_{n_{t},i}\right) \Big],
\end{align}
% \begin{align}
%     \notag \max_\theta \sum_{t=1}^T\sum_{i=1}^k \left[&\log p_\theta(s_{p_{t},i}|t, p_t, a_{p_{t},i})\right.+\\
%     &\left.\log p_\theta(s_{n_{t},i}|t, n_t, a_{n_{t},i})\right],
% \end{align}
% \begin{equation}
%     \min_\theta L(\theta)\triangleq -\frac{1}{l}\sum_{i=1}^{l} \log f_\theta(y|t, p_t, a_{p_{t}}),
% \end{equation}
% where $l$ denotes the length. Both positive $a_{p_{t}}$ and negative $a_{n_{t}}$ arguments are used for training. 
Compared to existing instruction-tuning datasets that mainly focus on covering a broad range of topics, we only utilize a limited number of topics while each topic containing $2\times k$ samples covering $k$ diverse arguments, $2$ opposite stances per argument, and a high-quality salient statement for each stance. As shown in the experiments, our dataset significantly improves the LLM controllability. 
% Different from most instruction-tuning data which contains a broad variety of topics, the overall amount of topics of ours are limited, but each of which contains $2 \times k$ different variants, equipping a supreme controllability of the tuned LLM.

\section{Evaluation Metrics}

We evaluate LLM's ability to generate statements for controversial topics on two orthogonal aspects on our hold-out test set: the \textbf{Response Quality} and \textbf{Controversy Controllability}. The Response Quality measures whether the LLM can generate helpful, relevant, accurate, and detailed statements for an instruction, which aligns with the common requirements for LLM's responses. However, the response quality fails to measure the extent to which LLM's response is stuck to the desired stances. As illustrated previously, existing LLM tends to generate average viewpoints endorsed by the majority and neglect the voice of the minority, which might be graded highly by existing judging methods. Thus we propose another aspect, noted as Controversy Controllability, which directly measures whether LLM's response is strictly stuck to the desired stances. 

\subsection{LLM Judge}

Considering the large number of test samples, we utilize GPT4 as the judge for evaluation, 
% Evaluating the quality of responses of LLMs, especially for open-domain queries, is a complex area of ongoing research. One of the main difficulties lies in establishing a clear ground truth for such evaluations. However, recent trends in the field have seen the use of LLMs themselves, such as GPT-4, to evaluate responses. 
which has become widely accepted, as noted in several studies \cite{touvron2023llama2, vicuna2023, dettmers2023qlora, liu2023geval, chiang-lee-2023-large}. Research has demonstrated that the evaluations made by GPT-4 align well with human judgments \cite{zheng2023judging, alpaca_eval}. 
% Thus in our study, to measure the individual quality of each statement for the controversial topic, we follow the previous work utilizing GPT4 as the judge. 
\footnote{Both the detailed prompts for Response Quality and Controversy Controllability can be found in Appendix \ref{appendix_prompt}.}

\noindent
\textbf{Response Quality}:

The evaluation of Response Quality follows \citet{chen2023alpagasus, cherry, Li2024SuperfilteringWD}, which involves a detailed rating system for the responses generated by the model. This system compares responses generated by two different LLMs on various dimensions, including helpfulness, relevance, accuracy, and level of detail. We also address the issue of positional bias in the LLM judge system, as discussed in the studies by \citet{ko-etal-2020-look, wang2023large} by presenting models' responses in two separate sequences for evaluation by the LLM judge. We then analyze the responses for each instruction by comparing them through a "Win-Tie-Loss" system. Then the win score will be calculated for better comparison: 
\begin{equation}
\text{Score} = \frac{n_\text{Win} - n_\text{Lose}}{n_\text{All}} + 1,
\end{equation}

\noindent
\textbf{Controversy Controllability}: 

For better illustration, we utilize \textbf{Positive/Negative} to illustrate the stance for a debate topic as shown in Figure \ref{structure}. A Positive stance supports the topic sentence while a Negative stance is against the topic sentence. There can be diverse arguments under each stance. By changing the stance and the argument in the input, a controllable LLM should generate detailed and strong statements supporting the given stance and the argument, even if they only represent the minority's point of view. 
We utilize the \textbf{Good/Bad} pair to illustrate the success or failure of the LLM in generating the controllable statements. For example, if the LLM is prompted to support a topic and it successfully does so, then it is a good one; if it generates responses against the topic, then it would be a bad one.

For the evaluation of Controversy Controllability, we prompt GPT4 to analyze the response with the given topic without letting it know the specific stance of this response and ask it to guess and provide the supporting versus opposing proportion of the above arguments to the given topic. 
This method serves as a relaxation that turns the original complex problem into a problem similar to sentiment analysis, which is perfectly under the control of the powerful GPT4 model. Ideally, in a good sample, the majority proportion should be $100\%$ and is the same as the real given stance. Otherwise, it means this response fails to strictly stick to the user's query. 
Then we further categorize all the responses into Good or Bad ones and the Positive Controversy Controllability score is defined as the ratio of good samples in all positive samples, while the Negative one is the ratio in all negative samples. 
The Overall Controversy Controllability is the average of the Positive and Negative. The higher score represents the more samples are strictly stuck to their given stance, representing a better Controversy Controllability. 

\subsection{Human Study}

To further compare the Controversy Controllability of our model and the baseline model, further human studies are conducted. Since there are $80$ topics in our test sets, each of which contains $6$ different arguments, resulting in a total of $480$ query-response pairs, making it infeasible to manually inspect all the samples. Moreover, we empirically find the number of bad examples is few due to the current strong instruction-following ability of current LLMs, thus it is also infeasible to inspect only the small random set of testing samples. 

To overcome this problem, we utilize an LLM-Human interactive inspection method. After utilizing LLM as the Judge for the Controversy Controllability evaluation, we select all the bad cases detected by GPT4, and then randomly sample some good cases to construct a new evaluation set with $100$ instruction-response pairs. Then human participants are queried to judge whether these responses are strictly stuck to their given stances. There are $3$ choices given, (1) Good, representing the response is strictly stuck to the given stance; (2) Bad, representing the response contains opposite content; (3) Tie, representing the response is hard to judge. We conduct this human study on both the baseline Vicuna 7B v1.5 and our \ours-7B.

\begin{table*}[!tbh]
\centering
\scalebox{0.82}{
\begin{tabular}{l|cccc|ccc}
\hline
 & \multicolumn{4}{c|}{\textbf{Response Quality} (model vs. baseline)} & \multicolumn{3}{c}{\textbf{Controversy Controllability}} \\
& Win$\uparrow$ & Tie & Loss$\downarrow$ & Win score$\uparrow$ & Positive $\uparrow$ & Negative $\uparrow$ & Overall $\uparrow$ \\\hline
\ours-7B (ours, baseline) & - & - & - & 1.00 & 0.958 & 0.979  & \textbf{0.969}  \\ 
\ours-13B (ours) & 43 & 101 & 16 & \textbf{1.17} & 0.950 & 0.946  & 0.948  \\ 
\hline
Alpaca 7B \cite{alpaca} & 2 & 1 & 157 & 0.03 & 0.938 & 0.883 & 0.910  \\ 
WizardLM 7B \cite{xu2023wizardlm} & 3 & 12 & 145 & 0.11 & 0.833 & 0.704  & 0.768  \\ 
WizardLM 13B V1.2 \cite{xu2023wizardlm} & 14 & 99 & 47 & 0.79 & 0.800 & 0.708  & 0.754  \\ 
Vicuna 7B v1.5 \cite{vicuna2023}& 6 & 19 & 135 & 0.19 & 0.900 & 0.796  & 0.848  \\ 
Vicuna 13B v1.5 \cite{vicuna2023} & 5 & 36 & 119 & 0.29 & 0.867 & 0.858  & 0.863 \\ 
LLaMA2 Chat 7B \cite{touvron2023llama2} & 1 & 17 & 142 & 0.12 & 0.196 & 0.429  & 0.313  \\ 
LLaMA2 Chat 13B \cite{touvron2023llama2} & 3 & 26 & 131 & 0.20 & 0.338 & 0.317  & 0.327  \\ 
Zephyr 7B Alpha \cite{tunstall2023zephyr}& 7 & 29 & 124 & 0.27 & 0.879 & 0.713  & 0.796  \\ 
Zephyr 7B Beta \cite{tunstall2023zephyr}& 12 & 84 & 64 & 0.67 & 0.942 & 0.733  & 0.838  \\ 
% ChatGPT & -     & -       & - & -     \\
\hline
\end{tabular}
}
\caption{
\textbf{Response Quality} (\ours-7B as the baseline) and \textbf{Controversy Controllability} of our models and other LLMs on generating statements for controversial topics. \ours archives the highest quality and controllability, indicating its effectiveness on generating controllable responses for controversial topics.
}
\label{tbl:main}
\end{table*}

\section{Experimental Result}

\subsection{Results on Controversial Controllability}

Table \ref{tbl:main} showcases our main evaluation results on both the Response Quality and Controversy Controllability on the hold-out test set. 

In the \textbf{Response Quality} section, we report the win-tie-loss statistics and corresponding win scores between other models and \ours-7B.
% , which is calculated as:
% \begin{equation}
% \text{Score} = \frac{n_\text{Win} - n_\text{Lose}}{n_\text{All}} + 1,
% \end{equation}
The overall ranking of different models on the Response Quality basically aligns with their performance on common instruction-following benchmarks. For example, the Alpaca has the lowest win score, and commonly believed better models have relatively higher response quality scores. 
% while the model growing better with respect to instruction-following ability, their response quality tends to be higher. 

However, when it comes to \textbf{Controversial Controllability}, which measures the extent to which LLM's responses are stuck to the given stances, the results are not directly correlated to the original ability of LLMs, which reveals an interesting but long-neglect phenomenon. Under this setting, LLaMA2 Chat models achieve the lowest controllability scores, reasonable due to their strongly constrained alignment. Given a controversial topic, they have a strong tendency to refuse to answer or to find a safe middle ground to avoid potential harm. Though this strong alignment potentially avoids the offensiveness, it also loses the possibility to speak for the diverse perspectives. On the contrary, the Alpaca model achieves the highest score on controllability, indicating that they can provide statements strongly stuck to the given stances while having the lowest response quality. However, the manual inspection further explains this phenomenon that it is because of the relatively low instruction-following ability, that Alpaca tends to repeat the given argument with only a little new content, thus leading to high controllability and low quality. 

According to the above analysis, we can see both of these two criteria play an important role in evaluating the LLM's ability to generate statements for controversial topics. As shown in the results, our \ours, achieves the highest scores on both aspects compared with existing models, indicating our model's ability to speak for the minority. 
This Controversial Controllability metric proposed by us provides another dimension to examine the capability of current LLMs, pushing forward the understanding of their limitation and capabilities.
% We provide another dimension to examine the capability of current LLMs, pushing forward the understanding of their limitation and capabilities.

In the \textbf{Human Study}, there are $100$ samples examined generated by the baseline Vicuna 7B v1.5 and our \ours-7B. 
For our model, $87/100$ samples are inspected as Good, $2/100$ as Ti.e., and $11/100$ as Bad samples. For the Vicuna model, $40/100$ samples are inspected as Good, $7/100$ as Ti.e., and $53/100$ as Bad samples. 
In this human study, we carefully examined all the GPT-4 labeled bad examples by human experts and observed a similar and high ratio of human-labeled bad samples within the GPT4-labeled bad samples. Moreover, we also examined a random set of 100 good samples labeled by GPT4 and almost all of them are indeed good samples for human experts.
The large discrepancy between Vicuna and our model further verifies our method and the high consistency between human evaluation and LLM evaluation verifies the effectiveness of our evaluation method. 

% 87 samples are manually inspected as Good, while only 40 samples are inspected as Good for Vicuna. Moreover, only 11 bad samples and 2 tie samples for our model while there are 53 bad samples and 11 tie samples 

\begin{table*}[!tbh]
\centering
\scalebox{0.8}{
\begin{tabular}{l|cccc|ccc}
\hline
& \multicolumn{4}{c|}{\textbf{Response Quality} ( vs. Vicuna 7B v1.5)} & \multicolumn{3}{c}{\textbf{Controversy Controllability}} \\
& Win$\uparrow$ & Tie & Loss$\downarrow$ & Win score$\uparrow$ & Positive $\uparrow$ & Negative $\uparrow$ & Overall $\uparrow$ \\\hline
ShareGPT (Vicuna 7B v1.5, baseline) & - & - & - & \textbf{1.00} & 0.900 & 0.796  & 0.848  \\ 
\hline
Topic Data without Debate (3 Arguments) & 118 & 27 & 15 & 1.64 & 0.946 & 0.813  & 0.879  \\ 
1-round Debate per Topic (3 Arguments) & 134 & 19 & 7 & 1.79 & 0.954 & 0.950  & 0.952  \\ 
2-round Debate per Topic (3 Arguments) & 135 & 19 & 6 & \textbf{1.81} & 0.958 & 0.979  & \textbf{0.969}  \\ 
3-round Debate per Topic (3 Arguments) & 135 & 17 & 8 & 1.79 & 0.967 & 0.963  & 0.965  \\ 
\hline
% Train with 1/2 arguments & 116 & 39 & 5 & 1.69 & 0.973 & 0.958  & 0.966  \\ 
1 Argument per Topic (2-round Debate) & 135 & 21 & 4 & \textbf{1.82} & 0.946 & 0.921 & 0.933  \\ 
% M + Topic with 2 Arguments (2-round) & 130 & 27 & 3 & 1.79 & 0.981 & 0.973 & 0.977  \\ 
3 Arguments per Topic (2-round Debate) & 135 & 19 & 6 & 1.81 & 0.958 & 0.979  & \textbf{0.969}  \\ 
% M + Topic with 4 Arguments (2-round) & 138 & 16 & 6 & \textbf{1.83} & 0.981 & 0.971 & 0.976  \\ 
5 Arguments per Topic (2-round Debate) & 135 & 18 & 7 & 1.80 & 0.933 & 0.933 & 0.933  \\ 
\hline
\end{tabular}
}
\caption{
\textbf{Ablation study} on the number of debate rounds and the number of arguments per (topic, stance). Response Quality (Vicuna 7B v1.5 as the baseline) and Controversy Controllability are reported. It verifies the optimality of the default setting, i.e., $2$-round debate and $3$ arguments per topic. 
% $\mathcal{M}$ stands for the LLaMA2-7B model which is used as the base model. 
% The upper section verifies the effectiveness of our data and the debate process. The lower section explores the optimal argument settings. 
}
\label{tbl:ablation}
\end{table*}

\subsection{Ablation studies}

In this section, ablation studies are conducted to verify the configuration of our method. All experiments are conducted on the LLaMA-7B model. 
% denoted as $\mathcal{M}$ for simplicity. 
During the comparison, Vicuna-7B v1.5 is utilized as the baseline model as it is trained with diverse ShareGPT data containing real human queries. The results are shown in Table \ref{tbl:ablation}. 

The upper section of the table showcases the experiments with different debate configurations, 3 arguments are utilized for each stance of a given topic by default. ``Topic Data without Debate'' represents the model trained directly with the training split of our controversial topics, whose response is generated from \textit{gpt-3.5-turbo-1106} without debate.  
% \clc{avoid using ChatGPT here, GPT-3.5-turbo could be a better name since GPT-4 now is renamed as ChatGPT-4}. 
We can observe clear improvements in both the Response Quality and Controversy Controllability, indicating a rise in the capability of sticking to the given stance, which directly proves the effectiveness of our collected data. 

``x-round Debate on each Topic'' represents the model trained with debate-augmented responses for training. From the results, we can observe that even a one-round debate can significantly improve our model's capability on both two metrics. During the debate, the involved agent is required to strictly stick to the given stance, otherwise will be rebuked by the opponent. Then after rebuttal, the agent is able to further refine its previous response. This debate process improves the responses to controversial topics in 2 aspects: 
1. This process is naturally an interactive refinement process, thus continuously polishing the response itself, guaranteeing good response quality, which is proved by \citet{du2023improving, liang2023encouraging}. 
% ~\clc{could consider moving the citations to the related works}
2. This debate process requires the agent to think of the potential opposing responses and answer them in advance, and this thinking pattern increases the controversy controllability, similar to \citet{mukherjee2023orca, mitra2023orca}, which also tries to distill thinking patterns to student models. 

In the upper section, it is observed that a 2-round debate is the optimal setting, and thus extensive experiments are conducted as shown in the lower part aiming to find the optimal number of arguments for each stance of the given topic. The 3-Argument setting marginally outperforms the other options, thus we continuously set it as our default setting.

\begin{table*}[!htbh]
\centering
\scalebox{0.82}{
\begin{tabular}{l|c|ccccc|c|c}
\hline
 & \textbf{Pair-Wise}& \multicolumn{5}{c|}{\textbf{Huggingface Open LLM Leaderboard}} & \textbf{Alpaca Eval} & \textbf{MT Bench} \\
& Win Score & \textbf{Average} & ARC & HellaSwag & MMLU & TruthfulQA & Win Rate & Score \\
\hline 
Vicuna 7B v1.5 & 1.000 & 57.95 & 53.24 & 77.39 & 50.82 & 50.33 &73.10 & 6.07 \\
+ 1-Arg (2-round) & 1.220 & \textbf{58.47} & 54.10 & 77.20 & 51.17 & 51.40 & \textbf{79.70} & 5.90 \\
+ 3-Arg (2-round) & \textbf{1.257} & 57.77 & 52.56 & 76.54 & 51.08 & 50.91 & 78.76 & \textbf{6.13} \\
% + 5 Arg (2-round) & 1.188 & 58.11 & 53.16 & 76.48 & 51.13 & 51.67 & - & - \\
\hline
% Alpaca-GPT4 & 1.000 & - & - & - & - & - & - & - \\
% + 1 Arg (2-round) & - & - & - & - & - & - & - & - \\
% + 3 Arg (2-round) & - & - & - & - & - & - & - & - \\
% \hline
WizardLM 7B  & 1.000 & 57.09 & 54.18 & 79.25 & 46.92 & 48.01 & 66.08 & 5.56 \\
+ 1-Arg (2-round) & \textbf{1.372} & \textbf{57.72} & 54.69 & 78.61 & 46.96 & 50.62 & \textbf{74.04} & 5.57 \\
+ 3-Arg (2-round) & 1.339 & 57.46 & 54.86 & 78.12 & 46.94 & 49.90 & 71.20 & \textbf{5.70} \\
\hline
% sRecycled WizardLM  & 1.000 & - & - & - & - & - & - & - \\
% + 1 Arg (2-round) & - & - & - & - & - & - & - & - \\
% + 3 Arg (2-round) & - & - & - & - & - & - & - & - \\
% \hline
\end{tabular}
}
\caption{
Evaluation of \ours-trained models on three widely used benchmarks and pairwise comparison with the baseline models.  
% instruction-following evaluation metrics where we further instruction tune the corresponding LLMs utilizing our debate-augmented controversial topic data. 
By using only $630$ topics, \ours achieves consistent improvements on two different base LLMs and different evaluation metrics. 
% with only $630$ topics used in total, indicating a strong catalytic effect of our method. 
}
\label{tbl:if}
\end{table*}

\subsection{Results on Instruction Following}

In addition to the main results on our hold-out test set, evaluating the ability to generate statements for controversial topics, we also propose that our method can improve the general instruction following the ability of LLMs. To verify this, we directly finetune the Vicuna 7B v1.5 and WizardLM 7B (based on LLaMA2) models using the debate-augmented training set, containing $630$ topics, without data from any other sources. Then we evaluate our model on $4$ different commonly used methods, including \textbf{Pair-Wise Comparison},\textbf{ Huggingface Open LLM Leaderboard}, \textbf{Alpaca Eval Leaderboard} and \textbf{MT Bench}. \footnote{The evaluation metrics will be introduced detailedly in the Appendix \ref{appendix_if}.}

As shown in Table \ref{tbl:if}, the model further trained with our data outperforms the baseline models on all of the $4$ different evaluation metrics on two different models. It is worth noting that only $630$ topics are utilized, indicating the neglectable new knowledge involved in the training, while it causes a consistent improvement in the general instruction-following ability. We believe this is because of the high-quality responses generated during the debate mechanism. After the debate, the responses contain detailed statements that are strongly aligned with the controllable queries, this strong alignment further catalyzes the instruction-following ability of the models \cite{Li2024SelectiveRS, xu2024survey}. 

% We hypothesize that it is because the debate-like thinking patterns are distilled into the base model, causing steady improvement. 

\section{Related Work}

\subsection{LLM Alignment}

Despite the advancements of the current LLM, a fundamental issue with LLMs is the disjunction between their training objectives (i.e., minimizing contextual word prediction error), and users' aspirations for models (i.e., interpret and execute instructions reliably~\citep{radford2019language, brown2020language, Fedus2021SwitchTS}). To reconcile this, recent NLP research efforts focus on empowering LLMs to understand instructions and to align with human expectations, i.e., Instruction Tuning \cite{ye-etal-2021-crossfit, wei2022finetuned, wang-etal-2022-super, du-etal-2022-glm, honovich-etal-2023-unnatural, alpaca, vicuna2023, liu2023mmc}.  

% Overvoew of Alignment 
Another significant challenge in developing language models is ensuring that their output is useful, accurate, and consistent with human ethical standards \cite{kenton2021alignment, weidinger2021ethical, askell2021general, wang2023aligning}. A common method to achieve this involves engaging human raters to evaluate and compare the outputs of these models \cite{bai2022training, NEURIPS2022_b1efde53, stiennon2020learning, ziegler2019fine}. This feedback is crucial for improving the model's effectiveness in various tasks such as following instructions and answering questions. Recently the feedback from AI \cite{bai2022constitutional, lee2023rlaif} also benefits the alignment of LLMs. In the case of large-scale models, this method has been shown to enhance performance on specialized datasets aimed at assessing model alignment.

\subsection{Debate between LLMs}

With the continuous revealing of the self-improving ability \cite{self-improve, madaan2023self, selfee2023, Li2023ReflectionTuningDR} of LLMs, a Multiagent Debate framework \cite{du2023improving, liang2023encouraging} is proposed to further improve the responses of LLMs. The motivation of these methods is to reach a consensus for a given instruction, thus always requiring an additional Judge \cite{liang2023encouraging}, Confidence Estimator \cite{chen2023reconcile} or Summarizer \cite{chan2023chateval}. 
How to effectively reach the consensus in the debate framework is non-trivial \cite{chen2023multiagent} and still under exploring. 
Moreover, this debate framework is further used in the evaluation of LLMs \cite{wang-etal-2023-chatgpt-defend, chan2023chateval}, and helps non-expert judges identify the truth \cite{michael2023debate}. 

\subsection{Debate Datasets}

The exploration of debate-related datasets NLP has yielded significant resources, each contributing uniquely to the advancement of debating systems. DebateSum \cite{roush-balaji-2020-debatesum} is a large-scale dataset that includes a rich collection of debate documents with high-quality arguments, facilitating a variety of NLP tasks, especially argument mining and summarization, while the direct debate topic is not provided. Change My View \cite{hidey-etal-2017-analyzing} focuses on the effectiveness of arguments in changing viewpoints and SOCIAL-CHEM101 \cite{forbes-etal-2020-social} focuses on social norms, both of which are not suitable for debate. Argument Reasoning Comprehension Task \cite{habernal-etal-2018-argument} focuses on identifying and reconstructing implicit warrants in arguments. Moreover, IBM Project Debater \footnote{\url{https://research.ibm.com/haifa/dept/vst/debating_data.shtml}} \cite{shnarch-etal-2020-unsupervised, ein2020corpus, levy-etal-2018-towards, shnarch-etal-2018-will, gleize-etal-2019-convinced, toledo-etal-2019-automatic} also leads to the creation of diverse NLP datasets spanning various categories.

\section{Conclusion and Future Work}

Our study raises the long-neglect issue of generating controllable responses towards controversial topics, not only describing and exemplifying but also building an evaluation pipeline for the assessment and a novel method to alleviate this problem.  
More specifically, our \ours, is a novel pipeline that enhances model controllability over diverse perspectives on controversial topics. We have curated the largest dataset of debate topics to date and introduced a new metric for measuring controllability. Our evaluations reveal that LLMs can be effectively fine-tuned to represent a broader spectrum of opinions, paving the way for more inclusive AI-generated discourse.

In our study, it is shown that controversial arguments and statements are beneficial for LLMs in further generating diverse and high-quality responses supporting different controversial topics. However, our current method mainly focuses on generating desired statements by utilizing LLMs, though effective, a better strategy would be directly collecting high-quality human-written responses. One potential source would be directly collecting statements directly from the debating websites. Another potential source would be reforming the existing high-quality DebateSum dataset \cite{roush-balaji-2020-debatesum}. As mentioned in the previous section, DebateSum has diverse and high-quality statements while lacking corresponding topics and stances, thus further human annotation is required. We believe that the further collecting and combining of human-written high-quality data can further improve LLM's controllability to generate responses for diverse controversial topics and thus help minorities to express their own opinions. 

\section{Ethical Concerns }

To address ethical concerns, the LLMs used for debate are required to be the ones trained by safety alignment so they cannot generate toxic content or content with ethical concerns. These LLMs are required to pass the safety test before being deployed for debate. We will further apply an output filter to the debate-generated statements to double-confirm the safety of the finetuning data before using them for training LLMs. This will address the ethical concerns and meanwhile preserve the controllability of the resulting LLM and its capability to speak for minority groups. In addition, improving LLM’s controllability reduces the uncertainty of LLM outputting unexpected harmful content so humans can more effectively enforce the safety constraints directly through the “controls” in the input.

\section*{Limitations}

The main limitation of this work is the lack of studies on the agent used for debate. 
This work only considers the setting where ChatGPTs are utilized as the debate agents to ensure the quality of the responses. 
However, it would be more interesting to know if this debate framework can be effectively utilized on the existing relatively weak LLMs.

\bibliography{custom}

\begin{thebibliography}{71}
\expandafter\ifx\csname natexlab\endcsname\relax\def\natexlab#1{#1}\fi

\bibitem[{Askell et~al.(2021)Askell, Bai, Chen, Drain, Ganguli, Henighan, Jones, Joseph, Mann, DasSarma, Elhage, Hatfield-Dodds, Hernandez, Kernion, Ndousse, Olsson, Amodei, Brown, Clark, McCandlish, Olah, and Kaplan}]{askell2021general}
Amanda Askell, Yuntao Bai, Anna Chen, Dawn Drain, Deep Ganguli, Tom Henighan, Andy Jones, Nicholas Joseph, Ben Mann, Nova DasSarma, Nelson Elhage, Zac Hatfield-Dodds, Danny Hernandez, Jackson Kernion, Kamal Ndousse, Catherine Olsson, Dario Amodei, Tom Brown, Jack Clark, Sam McCandlish, Chris Olah, and Jared Kaplan. 2021.
\newblock \href {http://arxiv.org/abs/2112.00861} {A general language assistant as a laboratory for alignment}.

\bibitem[{Bai et~al.(2022{\natexlab{a}})Bai, Jones, Ndousse, Askell, Chen, DasSarma, Drain, Fort, Ganguli, Henighan, Joseph, Kadavath, Kernion, Conerly, El-Showk, Elhage, Hatfield-Dodds, Hernandez, Hume, Johnston, Kravec, Lovitt, Nanda, Olsson, Amodei, Brown, Clark, McCandlish, Olah, Mann, and Kaplan}]{bai2022training}
Yuntao Bai, Andy Jones, Kamal Ndousse, Amanda Askell, Anna Chen, Nova DasSarma, Dawn Drain, Stanislav Fort, Deep Ganguli, Tom Henighan, Nicholas Joseph, Saurav Kadavath, Jackson Kernion, Tom Conerly, Sheer El-Showk, Nelson Elhage, Zac Hatfield-Dodds, Danny Hernandez, Tristan Hume, Scott Johnston, Shauna Kravec, Liane Lovitt, Neel Nanda, Catherine Olsson, Dario Amodei, Tom Brown, Jack Clark, Sam McCandlish, Chris Olah, Ben Mann, and Jared Kaplan. 2022{\natexlab{a}}.
\newblock \href {http://arxiv.org/abs/2204.05862} {Training a helpful and harmless assistant with reinforcement learning from human feedback}.

\bibitem[{Bai et~al.(2022{\natexlab{b}})Bai, Kadavath, Kundu, Askell, Kernion, Jones, Chen, Goldie, Mirhoseini, McKinnon, Chen, Olsson, Olah, Hernandez, Drain, Ganguli, Li, Tran-Johnson, Perez, Kerr, Mueller, Ladish, Landau, Ndousse, Lukosuite, Lovitt, Sellitto, Elhage, Schiefer, Mercado, DasSarma, Lasenby, Larson, Ringer, Johnston, Kravec, Showk, Fort, Lanham, Telleen-Lawton, Conerly, Henighan, Hume, Bowman, Hatfield-Dodds, Mann, Amodei, Joseph, McCandlish, Brown, and Kaplan}]{bai2022constitutional}
Yuntao Bai, Saurav Kadavath, Sandipan Kundu, Amanda Askell, Jackson Kernion, Andy Jones, Anna Chen, Anna Goldie, Azalia Mirhoseini, Cameron McKinnon, Carol Chen, Catherine Olsson, Christopher Olah, Danny Hernandez, Dawn Drain, Deep Ganguli, Dustin Li, Eli Tran-Johnson, Ethan Perez, Jamie Kerr, Jared Mueller, Jeffrey Ladish, Joshua Landau, Kamal Ndousse, Kamile Lukosuite, Liane Lovitt, Michael Sellitto, Nelson Elhage, Nicholas Schiefer, Noemi Mercado, Nova DasSarma, Robert Lasenby, Robin Larson, Sam Ringer, Scott Johnston, Shauna Kravec, Sheer~El Showk, Stanislav Fort, Tamera Lanham, Timothy Telleen-Lawton, Tom Conerly, Tom Henighan, Tristan Hume, Samuel~R. Bowman, Zac Hatfield-Dodds, Ben Mann, Dario Amodei, Nicholas Joseph, Sam McCandlish, Tom Brown, and Jared Kaplan. 2022{\natexlab{b}}.
\newblock \href {http://arxiv.org/abs/2212.08073} {Constitutional ai: Harmlessness from ai feedback}.

\bibitem[{Bakker et~al.(2022)Bakker, Chadwick, Sheahan, Tessler, Campbell-Gillingham, Balaguer, McAleese, Glaese, Aslanides, Botvinick, and Summerfield}]{bakker2022fine}
Michiel~A. Bakker, Martin~J Chadwick, Hannah Sheahan, Michael~Henry Tessler, Lucy Campbell-Gillingham, Jan Balaguer, Nat McAleese, Amelia Glaese, John Aslanides, Matthew Botvinick, and Christopher Summerfield. 2022.
\newblock \href {https://openreview.net/forum?id=G5ADoRKiTyJ} {Fine-tuning language models to find agreement among humans with diverse preferences}.
\newblock In \emph{Advances in Neural Information Processing Systems}.

\bibitem[{Brown et~al.(2020)Brown, Mann, Ryder, Subbiah, Kaplan, Dhariwal, Neelakantan, Shyam, Sastry, Askell, Agarwal, Herbert-Voss, Krueger, Henighan, Child, Ramesh, Ziegler, Wu, Winter, Hesse, Chen, Sigler, Litwin, Gray, Chess, Clark, Berner, McCandlish, Radford, Sutskever, and Amodei}]{brown2020language}
Tom Brown, Benjamin Mann, Nick Ryder, Melanie Subbiah, Jared~D Kaplan, Prafulla Dhariwal, Arvind Neelakantan, Pranav Shyam, Girish Sastry, Amanda Askell, Sandhini Agarwal, Ariel Herbert-Voss, Gretchen Krueger, Tom Henighan, Rewon Child, Aditya Ramesh, Daniel Ziegler, Jeffrey Wu, Clemens Winter, Chris Hesse, Mark Chen, Eric Sigler, Mateusz Litwin, Scott Gray, Benjamin Chess, Jack Clark, Christopher Berner, Sam McCandlish, Alec Radford, Ilya Sutskever, and Dario Amodei. 2020.
\newblock \href {https://proceedings.neurips.cc/paper_files/paper/2020/file/1457c0d6bfcb4967418bfb8ac142f64a-Paper.pdf} {Language models are few-shot learners}.
\newblock In \emph{Advances in Neural Information Processing Systems}, volume~33, pages 1877--1901. Curran Associates, Inc.

\bibitem[{Chan et~al.(2023)Chan, Chen, Su, Yu, Xue, Zhang, Fu, and Liu}]{chan2023chateval}
Chi-Min Chan, Weize Chen, Yusheng Su, Jianxuan Yu, Wei Xue, Shanghang Zhang, Jie Fu, and Zhiyuan Liu. 2023.
\newblock \href {http://arxiv.org/abs/2308.07201} {Chateval: Towards better llm-based evaluators through multi-agent debate}.

\bibitem[{Chen et~al.(2023{\natexlab{a}})Chen, Ji, Xu, and Zhao}]{chen2023multiagent}
Huaben Chen, Wenkang Ji, Lufeng Xu, and Shiyu Zhao. 2023{\natexlab{a}}.
\newblock \href {http://arxiv.org/abs/2310.20151} {Multi-agent consensus seeking via large language models}.

\bibitem[{Chen et~al.(2023{\natexlab{b}})Chen, Saha, and Bansal}]{chen2023reconcile}
Justin Chih-Yao Chen, Swarnadeep Saha, and Mohit Bansal. 2023{\natexlab{b}}.
\newblock \href {http://arxiv.org/abs/2309.13007} {Reconcile: Round-table conference improves reasoning via consensus among diverse llms}.

\bibitem[{Chen et~al.(2023{\natexlab{c}})Chen, Li, Yan, Wang, Gunaratna, Yadav, Tang, Srinivasan, Zhou, Huang, and Jin}]{chen2023alpagasus}
Lichang Chen, Shiyang Li, Jun Yan, Hai Wang, Kalpa Gunaratna, Vikas Yadav, Zheng Tang, Vijay Srinivasan, Tianyi Zhou, Heng Huang, and Hongxia Jin. 2023{\natexlab{c}}.
\newblock \href {http://arxiv.org/abs/2307.08701} {Alpagasus: Training a better alpaca with fewer data}.

\bibitem[{Chiang and Lee(2023)}]{chiang-lee-2023-large}
Cheng-Han Chiang and Hung-yi Lee. 2023.
\newblock \href {https://doi.org/10.18653/v1/2023.acl-long.870} {Can large language models be an alternative to human evaluations?}
\newblock In \emph{Proceedings of the 61st Annual Meeting of the Association for Computational Linguistics (Volume 1: Long Papers)}, pages 15607--15631, Toronto, Canada. Association for Computational Linguistics.

\bibitem[{Chiang et~al.(2023)Chiang, Li, Lin, Sheng, Wu, Zhang, Zheng, Zhuang, Zhuang, Gonzalez, Stoica, and Xing}]{vicuna2023}
Wei-Lin Chiang, Zhuohan Li, Zi~Lin, Ying Sheng, Zhanghao Wu, Hao Zhang, Lianmin Zheng, Siyuan Zhuang, Yonghao Zhuang, Joseph~E. Gonzalez, Ion Stoica, and Eric~P. Xing. 2023.
\newblock \href {https://lmsys.org/blog/2023-03-30-vicuna/} {Vicuna: An open-source chatbot impressing gpt-4 with 90\%* chatgpt quality}.

\bibitem[{Chowdhery et~al.(2022)Chowdhery, Narang, Devlin, Bosma, Mishra, Roberts, Barham, Chung, Sutton, Gehrmann, Schuh, Shi, Tsvyashchenko, Maynez, Rao, Barnes, Tay, Shazeer, Prabhakaran, Reif, Du, Hutchinson, Pope, Bradbury, Austin, Isard, Gur-Ari, Yin, Duke, Levskaya, Ghemawat, Dev, Michalewski, Garcia, Misra, Robinson, Fedus, Zhou, Ippolito, Luan, Lim, Zoph, Spiridonov, Sepassi, Dohan, Agrawal, Omernick, Dai, Pillai, Pellat, Lewkowycz, Moreira, Child, Polozov, Lee, Zhou, Wang, Saeta, Diaz, Firat, Catasta, Wei, Meier-Hellstern, Eck, Dean, Petrov, and Fiedel}]{Chowdhery2022PaLMSL}
Aakanksha Chowdhery, Sharan Narang, Jacob Devlin, Maarten Bosma, Gaurav Mishra, Adam Roberts, Paul Barham, Hyung~Won Chung, Charles Sutton, Sebastian Gehrmann, Parker Schuh, Kensen Shi, Sasha Tsvyashchenko, Joshua Maynez, Abhishek Rao, Parker Barnes, Yi~Tay, Noam Shazeer, Vinodkumar Prabhakaran, Emily Reif, Nan Du, Ben Hutchinson, Reiner Pope, James Bradbury, Jacob Austin, Michael Isard, Guy Gur-Ari, Pengcheng Yin, Toju Duke, Anselm Levskaya, Sanjay Ghemawat, Sunipa Dev, Henryk Michalewski, Xavier Garcia, Vedant Misra, Kevin Robinson, Liam Fedus, Denny Zhou, Daphne Ippolito, David Luan, Hyeontaek Lim, Barret Zoph, Alexander Spiridonov, Ryan Sepassi, David Dohan, Shivani Agrawal, Mark Omernick, Andrew~M. Dai, Thanumalayan~Sankaranarayana Pillai, Marie Pellat, Aitor Lewkowycz, Erica Moreira, Rewon Child, Oleksandr Polozov, Katherine Lee, Zongwei Zhou, Xuezhi Wang, Brennan Saeta, Mark Diaz, Orhan Firat, Michele Catasta, Jason Wei, Kathy Meier-Hellstern, Douglas Eck, Jeff Dean, Slav Petrov, and Noah Fiedel. 2022.
\newblock \href {http://arxiv.org/abs/2204.02311} {Palm: Scaling language modeling with pathways}.

\bibitem[{Clark et~al.(2018)Clark, Cowhey, Etzioni, Khot, Sabharwal, Schoenick, and Tafjord}]{clark2018think}
Peter Clark, Isaac Cowhey, Oren Etzioni, Tushar Khot, Ashish Sabharwal, Carissa Schoenick, and Oyvind Tafjord. 2018.
\newblock \href {http://arxiv.org/abs/1803.05457} {Think you have solved question answering? try arc, the ai2 reasoning challenge}.

\bibitem[{Dettmers et~al.(2023)Dettmers, Pagnoni, Holtzman, and Zettlemoyer}]{dettmers2023qlora}
Tim Dettmers, Artidoro Pagnoni, Ari Holtzman, and Luke Zettlemoyer. 2023.
\newblock \href {http://arxiv.org/abs/2305.14314} {Qlora: Efficient finetuning of quantized llms}.

\bibitem[{Ding and Ito(2023)}]{ding2023selfagreement}
Shiyao Ding and Takayuki Ito. 2023.
\newblock \href {http://arxiv.org/abs/2305.11460} {Self-agreement: A framework for fine-tuning language models to find agreement among diverse opinions}.

\bibitem[{Du et~al.(2023)Du, Li, Torralba, Tenenbaum, and Mordatch}]{du2023improving}
Yilun Du, Shuang Li, Antonio Torralba, Joshua~B. Tenenbaum, and Igor Mordatch. 2023.
\newblock \href {http://arxiv.org/abs/2305.14325} {Improving factuality and reasoning in language models through multiagent debate}.

\bibitem[{Du et~al.(2022)Du, Qian, Liu, Ding, Qiu, Yang, and Tang}]{du-etal-2022-glm}
Zhengxiao Du, Yujie Qian, Xiao Liu, Ming Ding, Jiezhong Qiu, Zhilin Yang, and Jie Tang. 2022.
\newblock \href {https://doi.org/10.18653/v1/2022.acl-long.26} {{GLM}: General language model pretraining with autoregressive blank infilling}.
\newblock In \emph{Proceedings of the 60th Annual Meeting of the Association for Computational Linguistics (Volume 1: Long Papers)}, pages 320--335, Dublin, Ireland. Association for Computational Linguistics.

\bibitem[{Dubois et~al.(2024)Dubois, Li, Taori, Zhang, Gulrajani, Ba, Guestrin, Liang, and Hashimoto}]{dubois2023alpacafarm}
Yann Dubois, Xuechen Li, Rohan Taori, Tianyi Zhang, Ishaan Gulrajani, Jimmy Ba, Carlos Guestrin, Percy Liang, and Tatsunori~B. Hashimoto. 2024.
\newblock \href {http://arxiv.org/abs/2305.14387} {Alpacafarm: A simulation framework for methods that learn from human feedback}.

\bibitem[{Ein{-}Dor et~al.(2020)Ein{-}Dor, Shnarch, Dankin, Halfon, Sznajder, Gera, Alzate, Gleize, Choshen, Hou, Bilu, Aharonov, and Slonim}]{ein2020corpus}
Liat Ein{-}Dor, Eyal Shnarch, Lena Dankin, Alon Halfon, Benjamin Sznajder, Ariel Gera, Carlos Alzate, Martin Gleize, Leshem Choshen, Yufang Hou, Yonatan Bilu, Ranit Aharonov, and Noam Slonim. 2020.
\newblock \href {https://doi.org/10.1609/AAAI.V34I05.6270} {Corpus wide argument mining - {A} working solution}.
\newblock In \emph{The Thirty-Fourth {AAAI} Conference on Artificial Intelligence, {AAAI} 2020, The Thirty-Second Innovative Applications of Artificial Intelligence Conference, {IAAI} 2020, The Tenth {AAAI} Symposium on Educational Advances in Artificial Intelligence, {EAAI} 2020, New York, NY, USA, February 7-12, 2020}, pages 7683--7691. {AAAI} Press.

\bibitem[{Fedus et~al.(2022)Fedus, Zoph, and Shazeer}]{Fedus2021SwitchTS}
William Fedus, Barret Zoph, and Noam Shazeer. 2022.
\newblock \href {http://jmlr.org/papers/v23/21-0998.html} {Switch transformers: Scaling to trillion parameter models with simple and efficient sparsity}.
\newblock \emph{J. Mach. Learn. Res.}, 23:120:1--120:39.

\bibitem[{Forbes et~al.(2020)Forbes, Hwang, Shwartz, Sap, and Choi}]{forbes-etal-2020-social}
Maxwell Forbes, Jena~D. Hwang, Vered Shwartz, Maarten Sap, and Yejin Choi. 2020.
\newblock \href {https://doi.org/10.18653/v1/2020.emnlp-main.48} {Social chemistry 101: Learning to reason about social and moral norms}.
\newblock In \emph{Proceedings of the 2020 Conference on Empirical Methods in Natural Language Processing (EMNLP)}, pages 653--670, Online. Association for Computational Linguistics.

\bibitem[{Gao et~al.(2021)Gao, Tow, Biderman, Black, DiPofi, Foster, Golding, Hsu, McDonell, Muennighoff, Phang, Reynolds, Tang, Thite, Wang, Wang, and Zou}]{eval-harness}
Leo Gao, Jonathan Tow, Stella Biderman, Sid Black, Anthony DiPofi, Charles Foster, Laurence Golding, Jeffrey Hsu, Kyle McDonell, Niklas Muennighoff, Jason Phang, Laria Reynolds, Eric Tang, Anish Thite, Ben Wang, Kevin Wang, and Andy Zou. 2021.
\newblock \href {https://doi.org/10.5281/zenodo.5371628} {A framework for few-shot language model evaluation}.

\bibitem[{Gleize et~al.(2019)Gleize, Shnarch, Choshen, Dankin, Moshkowich, Aharonov, and Slonim}]{gleize-etal-2019-convinced}
Martin Gleize, Eyal Shnarch, Leshem Choshen, Lena Dankin, Guy Moshkowich, Ranit Aharonov, and Noam Slonim. 2019.
\newblock \href {https://doi.org/10.18653/v1/P19-1093} {Are you convinced? choosing the more convincing evidence with a {S}iamese network}.
\newblock In \emph{Proceedings of the 57th Annual Meeting of the Association for Computational Linguistics}, pages 967--976, Florence, Italy. Association for Computational Linguistics.

\bibitem[{Gretz et~al.(2020)Gretz, Friedman, Cohen{-}Karlik, Toledo, Lahav, Aharonov, and Slonim}]{gretz2020large}
Shai Gretz, Roni Friedman, Edo Cohen{-}Karlik, Assaf Toledo, Dan Lahav, Ranit Aharonov, and Noam Slonim. 2020.
\newblock \href {https://doi.org/10.1609/AAAI.V34I05.6285} {A large-scale dataset for argument quality ranking: Construction and analysis}.
\newblock In \emph{The Thirty-Fourth {AAAI} Conference on Artificial Intelligence, {AAAI} 2020, The Thirty-Second Innovative Applications of Artificial Intelligence Conference, {IAAI} 2020, The Tenth {AAAI} Symposium on Educational Advances in Artificial Intelligence, {EAAI} 2020, New York, NY, USA, February 7-12, 2020}, pages 7805--7813. {AAAI} Press.

\bibitem[{Habernal et~al.(2018)Habernal, Wachsmuth, Gurevych, and Stein}]{habernal-etal-2018-argument}
Ivan Habernal, Henning Wachsmuth, Iryna Gurevych, and Benno Stein. 2018.
\newblock \href {https://doi.org/10.18653/v1/N18-1175} {The argument reasoning comprehension task: Identification and reconstruction of implicit warrants}.
\newblock In \emph{Proceedings of the 2018 Conference of the North {A}merican Chapter of the Association for Computational Linguistics: Human Language Technologies, Volume 1 (Long Papers)}, pages 1930--1940, New Orleans, Louisiana. Association for Computational Linguistics.

\bibitem[{Hendrycks et~al.(2021)Hendrycks, Burns, Basart, Zou, Mazeika, Song, and Steinhardt}]{hendrycks2021measuring}
Dan Hendrycks, Collin Burns, Steven Basart, Andy Zou, Mantas Mazeika, Dawn Song, and Jacob Steinhardt. 2021.
\newblock \href {https://openreview.net/forum?id=d7KBjmI3GmQ} {Measuring massive multitask language understanding}.
\newblock In \emph{International Conference on Learning Representations}.

\bibitem[{Hidey et~al.(2017)Hidey, Musi, Hwang, Muresan, and McKeown}]{hidey-etal-2017-analyzing}
Christopher Hidey, Elena Musi, Alyssa Hwang, Smaranda Muresan, and Kathy McKeown. 2017.
\newblock \href {https://doi.org/10.18653/v1/W17-5102} {Analyzing the semantic types of claims and premises in an online persuasive forum}.
\newblock In \emph{Proceedings of the 4th Workshop on Argument Mining}, pages 11--21, Copenhagen, Denmark. Association for Computational Linguistics.

\bibitem[{Honovich et~al.(2023)Honovich, Scialom, Levy, and Schick}]{honovich-etal-2023-unnatural}
Or~Honovich, Thomas Scialom, Omer Levy, and Timo Schick. 2023.
\newblock \href {https://aclanthology.org/2023.acl-long.806} {Unnatural instructions: Tuning language models with (almost) no human labor}.
\newblock In \emph{Proceedings of the 61st Annual Meeting of the Association for Computational Linguistics (Volume 1: Long Papers)}, pages 14409--14428, Toronto, Canada. Association for Computational Linguistics.

\bibitem[{Huang et~al.(2023)Huang, Gu, Hou, Wu, Wang, Yu, and Han}]{self-improve}
Jiaxin Huang, Shixiang Gu, Le~Hou, Yuexin Wu, Xuezhi Wang, Hongkun Yu, and Jiawei Han. 2023.
\newblock \href {https://doi.org/10.18653/v1/2023.emnlp-main.67} {Large language models can self-improve}.
\newblock In \emph{Proceedings of the 2023 Conference on Empirical Methods in Natural Language Processing}, pages 1051--1068, Singapore. Association for Computational Linguistics.

\bibitem[{Kenton et~al.(2021)Kenton, Everitt, Weidinger, Gabriel, Mikulik, and Irving}]{kenton2021alignment}
Zachary Kenton, Tom Everitt, Laura Weidinger, Iason Gabriel, Vladimir Mikulik, and Geoffrey Irving. 2021.
\newblock \href {http://arxiv.org/abs/2103.14659} {Alignment of language agents}.

\bibitem[{Ko et~al.(2020)Ko, Lee, Kim, Kim, and Kang}]{ko-etal-2020-look}
Miyoung Ko, Jinhyuk Lee, Hyunjae Kim, Gangwoo Kim, and Jaewoo Kang. 2020.
\newblock \href {https://doi.org/10.18653/v1/2020.emnlp-main.84} {Look at the first sentence: Position bias in question answering}.
\newblock In \emph{Proceedings of the 2020 Conference on Empirical Methods in Natural Language Processing (EMNLP)}, pages 1109--1121, Online. Association for Computational Linguistics.

\bibitem[{Lee et~al.(2023)Lee, Phatale, Mansoor, Mesnard, Ferret, Lu, Bishop, Hall, Carbune, Rastogi, and Prakash}]{lee2023rlaif}
Harrison Lee, Samrat Phatale, Hassan Mansoor, Thomas Mesnard, Johan Ferret, Kellie Lu, Colton Bishop, Ethan Hall, Victor Carbune, Abhinav Rastogi, and Sushant Prakash. 2023.
\newblock \href {http://arxiv.org/abs/2309.00267} {Rlaif: Scaling reinforcement learning from human feedback with ai feedback}.

\bibitem[{Levy et~al.(2018)Levy, Bogin, Gretz, Aharonov, and Slonim}]{levy-etal-2018-towards}
Ran Levy, Ben Bogin, Shai Gretz, Ranit Aharonov, and Noam Slonim. 2018.
\newblock \href {https://aclanthology.org/C18-1176} {Towards an argumentative content search engine using weak supervision}.
\newblock In \emph{Proceedings of the 27th International Conference on Computational Linguistics}, pages 2066--2081, Santa Fe, New Mexico, USA. Association for Computational Linguistics.

\bibitem[{Li et~al.(2024{\natexlab{a}})Li, Chen, Chen, He, Gu, and Zhou}]{Li2024SelectiveRS}
Ming Li, Lichang Chen, Jiuhai Chen, Shwai He, Jiuxiang Gu, and Tianyi Zhou. 2024{\natexlab{a}}.
\newblock \href {https://api.semanticscholar.org/CorpusID:267682220} {Selective reflection-tuning: Student-selected data recycling for llm instruction-tuning}.
\newblock \emph{ArXiv}, abs/2402.10110.

\bibitem[{Li et~al.(2023{\natexlab{a}})Li, Chen, Chen, He, and Zhou}]{Li2023ReflectionTuningDR}
Ming Li, Lichang Chen, Jiuhai Chen, Shwai He, and Tianyi Zhou. 2023{\natexlab{a}}.
\newblock \href {https://openreview.net/forum?id=xaqoZZqkPU} {Reflection-tuning: Recycling data for better instruction-tuning}.
\newblock In \emph{NeurIPS 2023 Workshop on Instruction Tuning and Instruction Following}.

\bibitem[{Li et~al.(2024{\natexlab{b}})Li, Zhang, He, Li, Zhao, Wang, Cheng, and Zhou}]{Li2024SuperfilteringWD}
Ming Li, Yong Zhang, Shwai He, Zhitao Li, Hongyu Zhao, Jianzong Wang, Ning Cheng, and Tianyi Zhou. 2024{\natexlab{b}}.
\newblock \href {https://api.semanticscholar.org/CorpusID:267365346} {Superfiltering: Weak-to-strong data filtering for fast instruction-tuning}.
\newblock \emph{ArXiv}, abs/2402.00530.

\bibitem[{Li et~al.(2023{\natexlab{b}})Li, Zhang, Li, Chen, Chen, Cheng, Wang, Zhou, and Xiao}]{cherry}
Ming Li, Yong Zhang, Zhitao Li, Jiuhai Chen, Lichang Chen, Ning Cheng, Jianzong Wang, Tianyi Zhou, and Jing Xiao. 2023{\natexlab{b}}.
\newblock \href {https://api.semanticscholar.org/CorpusID:261076515} {From quantity to quality: Boosting llm performance with self-guided data selection for instruction tuning}.
\newblock \emph{ArXiv}, abs/2308.12032.

\bibitem[{Li et~al.(2023{\natexlab{c}})Li, Zhang, Dubois, Taori, Gulrajani, Guestrin, Liang, and Hashimoto}]{alpaca_eval}
Xuechen Li, Tianyi Zhang, Yann Dubois, Rohan Taori, Ishaan Gulrajani, Carlos Guestrin, Percy Liang, and Tatsunori~B. Hashimoto. 2023{\natexlab{c}}.
\newblock Alpacaeval: An automatic evaluator of instruction-following models.
\newblock \url{https://github.com/tatsu-lab/alpaca_eval}.

\bibitem[{Liang et~al.(2023)Liang, He, Jiao, Wang, Wang, Wang, Yang, Tu, and Shi}]{liang2023encouraging}
Tian Liang, Zhiwei He, Wenxiang Jiao, Xing Wang, Yan Wang, Rui Wang, Yujiu Yang, Zhaopeng Tu, and Shuming Shi. 2023.
\newblock \href {http://arxiv.org/abs/2305.19118} {Encouraging divergent thinking in large language models through multi-agent debate}.

\bibitem[{Lin et~al.(2022)Lin, Hilton, and Evans}]{lin-etal-2022-truthfulqa}
Stephanie Lin, Jacob Hilton, and Owain Evans. 2022.
\newblock \href {https://doi.org/10.18653/v1/2022.acl-long.229} {{T}ruthful{QA}: Measuring how models mimic human falsehoods}.
\newblock In \emph{Proceedings of the 60th Annual Meeting of the Association for Computational Linguistics (Volume 1: Long Papers)}, pages 3214--3252, Dublin, Ireland. Association for Computational Linguistics.

\bibitem[{Liu et~al.(2023{\natexlab{a}})Liu, Wang, Yao, Chen, Song, Cho, Yacoob, and Yu}]{liu2023mmc}
Fuxiao Liu, Xiaoyang Wang, Wenlin Yao, Jianshu Chen, Kaiqiang Song, Sangwoo Cho, Yaser Yacoob, and Dong Yu. 2023{\natexlab{a}}.
\newblock \href {http://arxiv.org/abs/2311.10774} {Mmc: Advancing multimodal chart understanding with large-scale instruction tuning}.

\bibitem[{Liu et~al.(2023{\natexlab{b}})Liu, Iter, Xu, Wang, Xu, and Zhu}]{liu2023geval}
Yang Liu, Dan Iter, Yichong Xu, Shuohang Wang, Ruochen Xu, and Chenguang Zhu. 2023{\natexlab{b}}.
\newblock \href {http://arxiv.org/abs/2303.16634} {G-eval: Nlg evaluation using gpt-4 with better human alignment}.

\bibitem[{Madaan et~al.(2023)Madaan, Tandon, Gupta, Hallinan, Gao, Wiegreffe, Alon, Dziri, Prabhumoye, Yang, Gupta, Majumder, Hermann, Welleck, Yazdanbakhsh, and Clark}]{madaan2023self}
Aman Madaan, Niket Tandon, Prakhar Gupta, Skyler Hallinan, Luyu Gao, Sarah Wiegreffe, Uri Alon, Nouha Dziri, Shrimai Prabhumoye, Yiming Yang, Shashank Gupta, Bodhisattwa~Prasad Majumder, Katherine Hermann, Sean Welleck, Amir Yazdanbakhsh, and Peter Clark. 2023.
\newblock \href {http://arxiv.org/abs/2303.17651} {Self-refine: Iterative refinement with self-feedback}.

\bibitem[{Michael et~al.(2023)Michael, Mahdi, Rein, Petty, Dirani, Padmakumar, and Bowman}]{michael2023debate}
Julian Michael, Salsabila Mahdi, David Rein, Jackson Petty, Julien Dirani, Vishakh Padmakumar, and Samuel~R. Bowman. 2023.
\newblock \href {http://arxiv.org/abs/2311.08702} {Debate helps supervise unreliable experts}.

\bibitem[{Mitra et~al.(2023)Mitra, Corro, Mahajan, Codas, Simoes, Agarwal, Chen, Razdaibiedina, Jones, Aggarwal, Palangi, Zheng, Rosset, Khanpour, and Awadallah}]{mitra2023orca}
Arindam Mitra, Luciano~Del Corro, Shweti Mahajan, Andres Codas, Clarisse Simoes, Sahaj Agarwal, Xuxi Chen, Anastasia Razdaibiedina, Erik Jones, Kriti Aggarwal, Hamid Palangi, Guoqing Zheng, Corby Rosset, Hamed Khanpour, and Ahmed Awadallah. 2023.
\newblock \href {http://arxiv.org/abs/2311.11045} {Orca 2: Teaching small language models how to reason}.

\bibitem[{Mukherjee et~al.(2023)Mukherjee, Mitra, Jawahar, Agarwal, Palangi, and Awadallah}]{mukherjee2023orca}
Subhabrata Mukherjee, Arindam Mitra, Ganesh Jawahar, Sahaj Agarwal, Hamid Palangi, and Ahmed Awadallah. 2023.
\newblock \href {http://arxiv.org/abs/2306.02707} {Orca: Progressive learning from complex explanation traces of gpt-4}.

\bibitem[{Ouyang et~al.(2022)Ouyang, Wu, Jiang, Almeida, Wainwright, Mishkin, Zhang, Agarwal, Slama, Ray, Schulman, Hilton, Kelton, Miller, Simens, Askell, Welinder, Christiano, Leike, and Lowe}]{NEURIPS2022_b1efde53}
Long Ouyang, Jeffrey Wu, Xu~Jiang, Diogo Almeida, Carroll Wainwright, Pamela Mishkin, Chong Zhang, Sandhini Agarwal, Katarina Slama, Alex Ray, John Schulman, Jacob Hilton, Fraser Kelton, Luke Miller, Maddie Simens, Amanda Askell, Peter Welinder, Paul~F Christiano, Jan Leike, and Ryan Lowe. 2022.
\newblock \href {https://proceedings.neurips.cc/paper_files/paper/2022/file/b1efde53be364a73914f58805a001731-Paper-Conference.pdf} {Training language models to follow instructions with human feedback}.
\newblock In \emph{Advances in Neural Information Processing Systems}, volume~35, pages 27730--27744. Curran Associates, Inc.

\bibitem[{Papachristou et~al.(2024)Papachristou, Yang, and Hsu}]{papachristou2023leveraging}
Marios Papachristou, Longqi Yang, and Chin-Chia Hsu. 2024.
\newblock \href {http://arxiv.org/abs/2311.04928} {Leveraging large language models for collective decision-making}.

\bibitem[{Radford et~al.(2019)Radford, Wu, Child, Luan, Amodei, and Sutskever}]{radford2019language}
Alec Radford, Jeff Wu, Rewon Child, David Luan, Dario Amodei, and Ilya Sutskever. 2019.
\newblock \href {https://api.semanticscholar.org/CorpusID:160025533} {Language models are unsupervised multitask learners}.

\bibitem[{Roush and Balaji(2020)}]{roush-balaji-2020-debatesum}
Allen Roush and Arvind Balaji. 2020.
\newblock \href {https://aclanthology.org/2020.argmining-1.1} {{D}ebate{S}um: A large-scale argument mining and summarization dataset}.
\newblock In \emph{Proceedings of the 7th Workshop on Argument Mining}, pages 1--7, Online. Association for Computational Linguistics.

\bibitem[{Shnarch et~al.(2018)Shnarch, Alzate, Dankin, Gleize, Hou, Choshen, Aharonov, and Slonim}]{shnarch-etal-2018-will}
Eyal Shnarch, Carlos Alzate, Lena Dankin, Martin Gleize, Yufang Hou, Leshem Choshen, Ranit Aharonov, and Noam Slonim. 2018.
\newblock \href {https://doi.org/10.18653/v1/P18-2095} {Will it blend? blending weak and strong labeled data in a neural network for argumentation mining}.
\newblock In \emph{Proceedings of the 56th Annual Meeting of the Association for Computational Linguistics (Volume 2: Short Papers)}, pages 599--605, Melbourne, Australia. Association for Computational Linguistics.

\bibitem[{Shnarch et~al.(2020)Shnarch, Choshen, Moshkowich, Aharonov, and Slonim}]{shnarch-etal-2020-unsupervised}
Eyal Shnarch, Leshem Choshen, Guy Moshkowich, Ranit Aharonov, and Noam Slonim. 2020.
\newblock \href {https://doi.org/10.18653/v1/2020.findings-emnlp.243} {Unsupervised expressive rules provide explainability and assist human experts grasping new domains}.
\newblock In \emph{Findings of the Association for Computational Linguistics: EMNLP 2020}, pages 2678--2697, Online. Association for Computational Linguistics.

\bibitem[{Stiennon et~al.(2020)Stiennon, Ouyang, Wu, Ziegler, Lowe, Voss, Radford, Amodei, and Christiano}]{stiennon2020learning}
Nisan Stiennon, Long Ouyang, Jeffrey Wu, Daniel~M. Ziegler, Ryan Lowe, Chelsea Voss, Alec Radford, Dario Amodei, and Paul~F. Christiano. 2020.
\newblock \href {https://proceedings.neurips.cc/paper/2020/hash/1f89885d556929e98d3ef9b86448f951-Abstract.html} {Learning to summarize with human feedback}.
\newblock In \emph{Advances in Neural Information Processing Systems 33: Annual Conference on Neural Information Processing Systems 2020, NeurIPS 2020, December 6-12, 2020, virtual}.

\bibitem[{Taori et~al.(2023)Taori, Gulrajani, Zhang, Dubois, Li, Guestrin, Liang, and Hashimoto}]{alpaca}
Rohan Taori, Ishaan Gulrajani, Tianyi Zhang, Yann Dubois, Xuechen Li, Carlos Guestrin, Percy Liang, and Tatsunori~B. Hashimoto. 2023.
\newblock Stanford alpaca: An instruction-following llama model.
\newblock \url{https://github.com/tatsu-lab/stanford_alpaca}.

\bibitem[{Toledo et~al.(2019)Toledo, Gretz, Cohen-Karlik, Friedman, Venezian, Lahav, Jacovi, Aharonov, and Slonim}]{toledo-etal-2019-automatic}
Assaf Toledo, Shai Gretz, Edo Cohen-Karlik, Roni Friedman, Elad Venezian, Dan Lahav, Michal Jacovi, Ranit Aharonov, and Noam Slonim. 2019.
\newblock \href {https://doi.org/10.18653/v1/D19-1564} {Automatic argument quality assessment - new datasets and methods}.
\newblock In \emph{Proceedings of the 2019 Conference on Empirical Methods in Natural Language Processing and the 9th International Joint Conference on Natural Language Processing (EMNLP-IJCNLP)}, pages 5625--5635, Hong Kong, China. Association for Computational Linguistics.

\bibitem[{Touvron et~al.(2023{\natexlab{a}})Touvron, Lavril, Izacard, Martinet, Lachaux, Lacroix, Rozière, Goyal, Hambro, Azhar, Rodriguez, Joulin, Grave, and Lample}]{Touvron2023LLaMAOA}
Hugo Touvron, Thibaut Lavril, Gautier Izacard, Xavier Martinet, Marie-Anne Lachaux, Timothée Lacroix, Baptiste Rozière, Naman Goyal, Eric Hambro, Faisal Azhar, Aurelien Rodriguez, Armand Joulin, Edouard Grave, and Guillaume Lample. 2023{\natexlab{a}}.
\newblock \href {http://arxiv.org/abs/2302.13971} {Llama: Open and efficient foundation language models}.

\bibitem[{Touvron et~al.(2023{\natexlab{b}})Touvron, Martin, Stone, Albert, Almahairi, Babaei, Bashlykov, Batra, Bhargava, Bhosale, Bikel, Blecher, Ferrer, Chen, Cucurull, Esiobu, Fernandes, Fu, Fu, Fuller, Gao, Goswami, Goyal, Hartshorn, Hosseini, Hou, Inan, Kardas, Kerkez, Khabsa, Kloumann, Korenev, Koura, Lachaux, Lavril, Lee, Liskovich, Lu, Mao, Martinet, Mihaylov, Mishra, Molybog, Nie, Poulton, Reizenstein, Rungta, Saladi, Schelten, Silva, Smith, Subramanian, Tan, Tang, Taylor, Williams, Kuan, Xu, Yan, Zarov, Zhang, Fan, Kambadur, Narang, Rodriguez, Stojnic, Edunov, and Scialom}]{touvron2023llama2}
Hugo Touvron, Louis Martin, Kevin Stone, Peter Albert, Amjad Almahairi, Yasmine Babaei, Nikolay Bashlykov, Soumya Batra, Prajjwal Bhargava, Shruti Bhosale, Dan Bikel, Lukas Blecher, Cristian~Canton Ferrer, Moya Chen, Guillem Cucurull, David Esiobu, Jude Fernandes, Jeremy Fu, Wenyin Fu, Brian Fuller, Cynthia Gao, Vedanuj Goswami, Naman Goyal, Anthony Hartshorn, Saghar Hosseini, Rui Hou, Hakan Inan, Marcin Kardas, Viktor Kerkez, Madian Khabsa, Isabel Kloumann, Artem Korenev, Punit~Singh Koura, Marie-Anne Lachaux, Thibaut Lavril, Jenya Lee, Diana Liskovich, Yinghai Lu, Yuning Mao, Xavier Martinet, Todor Mihaylov, Pushkar Mishra, Igor Molybog, Yixin Nie, Andrew Poulton, Jeremy Reizenstein, Rashi Rungta, Kalyan Saladi, Alan Schelten, Ruan Silva, Eric~Michael Smith, Ranjan Subramanian, Xiaoqing~Ellen Tan, Binh Tang, Ross Taylor, Adina Williams, Jian~Xiang Kuan, Puxin Xu, Zheng Yan, Iliyan Zarov, Yuchen Zhang, Angela Fan, Melanie Kambadur, Sharan Narang, Aurelien Rodriguez, Robert Stojnic, Sergey Edunov, and Thomas
  Scialom. 2023{\natexlab{b}}.
\newblock \href {http://arxiv.org/abs/2307.09288} {Llama 2: Open foundation and fine-tuned chat models}.

\bibitem[{Tunstall et~al.(2023)Tunstall, Beeching, Lambert, Rajani, Rasul, Belkada, Huang, von Werra, Fourrier, Habib, Sarrazin, Sanseviero, Rush, and Wolf}]{tunstall2023zephyr}
Lewis Tunstall, Edward Beeching, Nathan Lambert, Nazneen Rajani, Kashif Rasul, Younes Belkada, Shengyi Huang, Leandro von Werra, Clémentine Fourrier, Nathan Habib, Nathan Sarrazin, Omar Sanseviero, Alexander~M. Rush, and Thomas Wolf. 2023.
\newblock \href {http://arxiv.org/abs/2310.16944} {Zephyr: Direct distillation of lm alignment}.

\bibitem[{Wang et~al.(2023{\natexlab{a}})Wang, Yue, and Sun}]{wang-etal-2023-chatgpt-defend}
Boshi Wang, Xiang Yue, and Huan Sun. 2023{\natexlab{a}}.
\newblock \href {https://doi.org/10.18653/v1/2023.findings-emnlp.795} {Can {C}hat{GPT} defend its belief in truth? evaluating {LLM} reasoning via debate}.
\newblock In \emph{Findings of the Association for Computational Linguistics: EMNLP 2023}, pages 11865--11881, Singapore. Association for Computational Linguistics.

\bibitem[{Wang et~al.(2023{\natexlab{b}})Wang, Li, Chen, Zhu, Lin, Cao, Liu, Liu, and Sui}]{wang2023large}
Peiyi Wang, Lei Li, Liang Chen, Dawei Zhu, Binghuai Lin, Yunbo Cao, Qi~Liu, Tianyu Liu, and Zhifang Sui. 2023{\natexlab{b}}.
\newblock \href {http://arxiv.org/abs/2305.17926} {Large language models are not fair evaluators}.

\bibitem[{Wang et~al.(2022)Wang, Mishra, Alipoormolabashi, Kordi, Mirzaei, Naik, Ashok, Dhanasekaran, Arunkumar, Stap, Pathak, Karamanolakis, Lai, Purohit, Mondal, Anderson, Kuznia, Doshi, Pal, Patel, Moradshahi, Parmar, Purohit, Varshney, Kaza, Verma, Puri, Karia, Doshi, Sampat, Mishra, Reddy~A, Patro, Dixit, and Shen}]{wang-etal-2022-super}
Yizhong Wang, Swaroop Mishra, Pegah Alipoormolabashi, Yeganeh Kordi, Amirreza Mirzaei, Atharva Naik, Arjun Ashok, Arut~Selvan Dhanasekaran, Anjana Arunkumar, David Stap, Eshaan Pathak, Giannis Karamanolakis, Haizhi Lai, Ishan Purohit, Ishani Mondal, Jacob Anderson, Kirby Kuznia, Krima Doshi, Kuntal~Kumar Pal, Maitreya Patel, Mehrad Moradshahi, Mihir Parmar, Mirali Purohit, Neeraj Varshney, Phani~Rohitha Kaza, Pulkit Verma, Ravsehaj~Singh Puri, Rushang Karia, Savan Doshi, Shailaja~Keyur Sampat, Siddhartha Mishra, Sujan Reddy~A, Sumanta Patro, Tanay Dixit, and Xudong Shen. 2022.
\newblock \href {https://aclanthology.org/2022.emnlp-main.340} {Super-{N}atural{I}nstructions: Generalization via declarative instructions on 1600+ {NLP} tasks}.
\newblock In \emph{Proceedings of the 2022 Conference on Empirical Methods in Natural Language Processing}, pages 5085--5109, Abu Dhabi, United Arab Emirates. Association for Computational Linguistics.

\bibitem[{Wang et~al.(2023{\natexlab{c}})Wang, Zhong, Li, Mi, Zeng, Huang, Shang, Jiang, and Liu}]{wang2023aligning}
Yufei Wang, Wanjun Zhong, Liangyou Li, Fei Mi, Xingshan Zeng, Wenyong Huang, Lifeng Shang, Xin Jiang, and Qun Liu. 2023{\natexlab{c}}.
\newblock \href {http://arxiv.org/abs/2307.12966} {Aligning large language models with human: A survey}.

\bibitem[{Wei et~al.(2022)Wei, Bosma, Zhao, Guu, Yu, Lester, Du, Dai, and Le}]{wei2022finetuned}
Jason Wei, Maarten Bosma, Vincent Zhao, Kelvin Guu, Adams~Wei Yu, Brian Lester, Nan Du, Andrew~M. Dai, and Quoc~V Le. 2022.
\newblock \href {https://openreview.net/forum?id=gEZrGCozdqR} {Finetuned language models are zero-shot learners}.
\newblock In \emph{International Conference on Learning Representations}.

\bibitem[{Weidinger et~al.(2021)Weidinger, Mellor, Rauh, Griffin, Uesato, Huang, Cheng, Glaese, Balle, Kasirzadeh, Kenton, Brown, Hawkins, Stepleton, Biles, Birhane, Haas, Rimell, Hendricks, Isaac, Legassick, Irving, and Gabriel}]{weidinger2021ethical}
Laura Weidinger, John Mellor, Maribeth Rauh, Conor Griffin, Jonathan Uesato, Po-Sen Huang, Myra Cheng, Mia Glaese, Borja Balle, Atoosa Kasirzadeh, Zac Kenton, Sasha Brown, Will Hawkins, Tom Stepleton, Courtney Biles, Abeba Birhane, Julia Haas, Laura Rimell, Lisa~Anne Hendricks, William Isaac, Sean Legassick, Geoffrey Irving, and Iason Gabriel. 2021.
\newblock \href {http://arxiv.org/abs/2112.04359} {Ethical and social risks of harm from language models}.

\bibitem[{Xu et~al.(2023)Xu, Sun, Zheng, Geng, Zhao, Feng, Tao, and Jiang}]{xu2023wizardlm}
Can Xu, Qingfeng Sun, Kai Zheng, Xiubo Geng, Pu~Zhao, Jiazhan Feng, Chongyang Tao, and Daxin Jiang. 2023.
\newblock \href {http://arxiv.org/abs/2304.12244} {Wizardlm: Empowering large language models to follow complex instructions}.

\bibitem[{Xu et~al.(2024)Xu, Li, Tao, Shen, Cheng, Li, Xu, Tao, and Zhou}]{xu2024survey}
Xiaohan Xu, Ming Li, Chongyang Tao, Tao Shen, Reynold Cheng, Jinyang Li, Can Xu, Dacheng Tao, and Tianyi Zhou. 2024.
\newblock \href {http://arxiv.org/abs/2402.13116} {A survey on knowledge distillation of large language models}.

\bibitem[{Ye et~al.(2021)Ye, Lin, and Ren}]{ye-etal-2021-crossfit}
Qinyuan Ye, Bill~Yuchen Lin, and Xiang Ren. 2021.
\newblock \href {https://doi.org/10.18653/v1/2021.emnlp-main.572} {{C}ross{F}it: A few-shot learning challenge for cross-task generalization in {NLP}}.
\newblock In \emph{Proceedings of the 2021 Conference on Empirical Methods in Natural Language Processing}, pages 7163--7189, Online and Punta Cana, Dominican Republic. Association for Computational Linguistics.

\bibitem[{Ye et~al.(2023)Ye, Jo, Kim, Kim, Hwang, and Seo}]{selfee2023}
Seonghyeon Ye, Yongrae Jo, Doyoung Kim, Sungdong Kim, Hyeonbin Hwang, and Minjoon Seo. 2023.
\newblock \href {https://kaistai.github.io/SelFee/} {Selfee: Iterative self-revising llm empowered by self-feedback generation}.
\newblock Blog post.

\bibitem[{Zellers et~al.(2019)Zellers, Holtzman, Bisk, Farhadi, and Choi}]{zellers-etal-2019-hellaswag}
Rowan Zellers, Ari Holtzman, Yonatan Bisk, Ali Farhadi, and Yejin Choi. 2019.
\newblock \href {https://doi.org/10.18653/v1/P19-1472} {{H}ella{S}wag: Can a machine really finish your sentence?}
\newblock In \emph{Proceedings of the 57th Annual Meeting of the Association for Computational Linguistics}, pages 4791--4800, Florence, Italy. Association for Computational Linguistics.

\bibitem[{Zheng et~al.(2023)Zheng, Chiang, Sheng, Zhuang, Wu, Zhuang, Lin, Li, Li, Xing, Zhang, Gonzalez, and Stoica}]{zheng2023judging}
Lianmin Zheng, Wei-Lin Chiang, Ying Sheng, Siyuan Zhuang, Zhanghao Wu, Yonghao Zhuang, Zi~Lin, Zhuohan Li, Dacheng Li, Eric.~P Xing, Hao Zhang, Joseph~E. Gonzalez, and Ion Stoica. 2023.
\newblock \href {http://arxiv.org/abs/2306.05685} {Judging llm-as-a-judge with mt-bench and chatbot arena}.

\bibitem[{Ziegler et~al.(2019)Ziegler, Stiennon, Wu, Brown, Radford, Amodei, Christiano, and Irving}]{ziegler2019fine}
Daniel~M. Ziegler, Nisan Stiennon, Jeffrey Wu, Tom~B. Brown, Alec Radford, Dario Amodei, Paul~F. Christiano, and Geoffrey Irving. 2019.
\newblock \href {http://arxiv.org/abs/1909.08593} {Fine-tuning language models from human preferences}.
\newblock \emph{CoRR}, abs/1909.08593.

\end{thebibliography}
% \bibliography{custom}

\appendix

\clearpage
\section{Prompt for Evaluation}
\label{appendix_prompt}

The prompts for evaluating the \textbf{Response Quality} and \textbf{Controversy Controllability} are shown in Figure \ref{prompt_1} and Figure \ref{prompt_2}, respectively.

\begin{figure}[h]
  \centering
  \parbox{0.48\textwidth}{
        \rule{0.48\textwidth}{1.5pt} % The line
        Prompt for Response Quality \\
        \rule{0.48\textwidth}{0.8pt} % The line
        \textbf{System Prompt} \\
        You are a helpful and precise assistant for checking the quality of the answer. \\

        \textbf{User Prompt} \\
        \text{[Question]}\\
        \textit{Question}\\
        \text{[The Start of Assistant 2's Answer]}\\
        \textit{Answer 2}\\
        \text{[The End of Assistant 2's Answer]}\\
        \text{[The Start of Assistant 2's Answer]}\\
        \textit{Answer 2}\\
        \text{[The End of Assistant 2's Answer]}\\

        We would like to request your feedback on the performance of two AI assistants in response to the user question displayed above. \\
        Please rate the helpfulness, relevance, accuracy, level of details of their responses. Each assistant receives an overall score on a scale of 1 to 10, where a higher score indicates better overall performance. \\
        Please first output a single line containing only two values indicating the scores for Assistant 1 and 2, respectively. The two scores are separated by a space. In the subsequent line, please provide a comprehensive explanation of your evaluation, avoiding any potential bias and ensuring that the order in which the responses were presented does not affect your judgment.

        \rule{0.48\textwidth}{0.8pt} % The line

  }
\caption{
The prompt to evaluate the Response Quality.  
} 
\label{prompt_1} 
\end{figure}

\begin{figure}[h]
  \centering
  \parbox{0.48\textwidth}{
        \rule{0.48\textwidth}{1.5pt} % The line
        Prompt for Controversy Controllability \\
        \rule{0.48\textwidth}{0.8pt} % The line
        \textbf{System Prompt} \\
        You are a helpful and precise assistant for checking the potential bias of arguments. \\

        \textbf{User Prompt} \\
        \text{[Topic]}\\
        \textit{Topic}\\
        \text{[The Start of Arguments]}\\
        \textit{Arguments}\\
        \text{[The End of Arguments]}\\

        We would like to request your feedback on the supporting versus opposing proportion of the above arguments to the given topic. Sometimes the overall arguments show a tendency, but some of the sentences indicate a different tendency. So please check the arguments seriously and answer how many percent of the arguments tend to support the topic, and how many percent of the arguments tend to oppose the topic. The scale is from 0 to 100. \\
        Please first output a single line containing only two values indicating the percentage of supporting and opposing proportions, respectively. The two scores are separated by a space. In the subsequent line, please provide explanations of your evaluation, avoiding any potential bias from your opinion of the topic.

        \rule{0.48\textwidth}{0.8pt} % The line

  }
\caption{
The prompt to evaluate the Controversy Controllability.  
} 
\label{prompt_2} 
\end{figure}

\clearpage
\section{Implementation Details}
\label{implementation}

In the debate process, we utilize \textit{gpt-3.5-turbo-1106} 
% \clc{specify the version of the model here or defer it to Appendix} 
as the default debating agent and the number of debating rounds is set to $2$ by default. For each stance on the topic, $3$ arguments are used for the process and for the training. We train our model based on both initial pretrained LLaMA2 \cite{touvron2023llama2} and Vicuna 7B v1.5 \cite{vicuna2023}. For the LLaMA2-based model, the learning rate is set to $2\times10^{-5}$ while $1\times10^{-5}$ for Vicuna-based model. The batch size is $128$, steer the training across $3$ epochs with a max length of $2048$. The warmup rate is set to $0.03$.

\clearpage
\section{Instruction-Following Evaluation}
\label{appendix_if}

\subsection{Pair-wise Comparison}

The setting and the prompt for the pair-wise comparison are the same as the evaluation for Response Quality as shown in Figure \ref{prompt_1}. The comparison is conducted on the WizardLM dataset \cite{xu2023wizardlm}, which contains 218 unique instructions, by utilizing GPT4 as the judge. 
The detailed comparison results are shown in the Table \ref{tbl:pair_result}

\begin{table}[!tbh]
\centering
\scalebox{0.9}{
\begin{tabular}{lcccc}
\hline
& Win & Tie & Lose & Win Rate\\\hline
Vicuna 7B v1.5      & -    & -       & -  & 1.000  \\
+ 1 Arg (2-round)   & 98     & 70 & 50       & 1.220   \\
+ 3 Arg (2-round)   & 99     & 76 & 43       & 1.257   \\
\hline
WizardLM 7B         & -    & -       & -  & 1.000  \\
+ 1 Arg (2-round)  & 119     & 61       & 38  & 1.372   \\
+ 3 Arg (2-round)  & 111     & 70       & 37  & 1.339   \\
\hline
\end{tabular}
}
\caption{
The pair-wise comparison between the debate-augmented model with the baseline models. 
}
\label{tbl:pair_result}
% \vspace{-2mm}
\end{table}

\subsection{Open LLM Leaderboard}

The Hugging Face Open LLM Leaderboard represents a cutting-edge initiative designed to showcase the performance of various LLMs across a wide array of benchmarks \cite{eval-harness}. It functions as a comprehensive and transparent platform where researchers and developers can compare the capabilities of different models based on standardized testing criteria. This leaderboard not only facilitates an objective evaluation of models in terms of natural language understanding, generation, and other AI tasks but also encourages the development of more efficient, accurate, and versatile language models. It focuses on $4$ pivotal benchmarks: ARC \cite{clark2018think}, HellaSwag \cite{zellers-etal-2019-hellaswag}, MMLU \cite{hendrycks2021measuring}, and TruthfulQA \cite{lin-etal-2022-truthfulqa}. 

\subsection{Alapca Eval Leaderboard}

The AlpacaEval Leaderboard provides a specialized platform for the automatic evaluation of Large Language Models (LLMs) using the AlpacaFarm evaluation dataset, as outlined in \citet{dubois2023alpacafarm}. This system offers an efficient and reliable method for assessing LLMs based on their ability to follow general user commands. Comparing model outputs with standard responses provided by Davinci003 ensures a comprehensive analysis. The system's effectiveness is highlighted by its strong correlation with human expert judgments, showcasing its accuracy in mirroring real-world expectations and the models' adherence to precise user instructions.

\subsection{MT-bench}

MT-Bench, the Multi-turn Benchmark proposed by \citet{vicuna2023}, serves as a rigorous framework for evaluating the conversational prowess of LLMs. It aims to measure how well these models can maintain coherent, informative, and engaging dialogue over multiple turns of conversation. This benchmark tests models on their ability to follow instructions and flow naturally in conversations, making it a crucial tool for assessing their performance in realistic dialogue scenarios. By focusing on the dynamic aspects of conversation, MT-Bench addresses a critical need in the AI community for benchmarks that can accurately reflect the capabilities of LLMs in engaging with users in a manner that mimics human conversation. 

% \clearpage
% \section{Controversy Controllability Examples}
% \label{appendix_cc_result}

% \clearpage
% \section{Instruction-Following Examples}
% \label{appendix_if_result}

\end{document}